\documentclass{article}

\usepackage[nonatbib,final]{neurips_2020}

\usepackage[utf8]{inputenc} 
\usepackage[T1]{fontenc}    
\usepackage{hyperref}       
\usepackage{url}            
\usepackage{booktabs}       
\usepackage{amsfonts}       
\usepackage{amsthm}
\usepackage{amssymb}
\usepackage{amsmath}
\usepackage{stackengine}
\usepackage{graphicx}
\usepackage{float}
\usepackage{biblatex}

\bibliography{bib/own,bib/all,bib/jonas}

\usepackage{textcomp}

\usepackage{nicefrac}       
\usepackage{microtype}      

\usepackage{xcolor}

\newtheorem{theorem}{Theorem}

\newcommand{\tabincell}[2]{\begin{tabular}{@{}#1@{}}#2\end{tabular}}

\title{Stochastic Normalizing Flows}

\author{%
  Hao Wu\\
  Tongji University\\Shanghai, P.R. China\\
  \texttt{wwtian@gmail.com} \\
  \And
  Jonas Köhler\\
  FU Berlin\\Berlin, Germany\\
  \texttt{jonas.koehler@fu-berlin.de} \\
  \And
  Frank Noé\\
  FU Berlin\\Berlin, Germany\\
  \texttt{frank.noe@fu-berlin.de} \\
}

\begin{document}

\maketitle

\begin{abstract}


The sampling of probability distributions specified up to a normalization constant is an important problem in both machine learning and statistical mechanics.
%
%
While classical stochastic sampling methods such as Markov Chain Monte Carlo (MCMC) or Langevin Dynamics (LD) can suffer from slow mixing times there is a growing interest in using normalizing flows in order to learn the transformation of a simple prior distribution to the given target distribution.
Here we propose a generalized and combined approach to sample target densities: Stochastic Normalizing Flows (SNF) – an arbitrary sequence of deterministic invertible functions and stochastic sampling blocks. 
We show that stochasticity overcomes expressivity limitations of normalizing flows resulting from the invertibility constraint, whereas trainable transformations between sampling steps improve efficiency of pure MCMC/LD along the flow.
%
%
By invoking ideas from non-equilibrium statistical mechanics we derive an efficient training procedure by which both the sampler's and the flow's parameters can be optimized end-to-end, and by which we can compute exact importance weights without having to marginalize out the randomness of the stochastic blocks.
We illustrate the representational power, sampling efficiency and asymptotic correctness of SNFs on several benchmarks including applications to sampling molecular systems in equilibrium.

\end{abstract}

\section{Introduction}



A common problem in machine learning and statistics with important applications in physics
is the generation of asymptotically unbiased samples from 
a target distribution defined up to a normalization constant by means of an
energy model $u(\mathbf{x})$:
\begin{equation}
\mu_X(\mathbf{x}) \propto \exp(-u(\mathbf{x})).
\end{equation}
Sampling of such unnormalized distributions is often done with
Markov Chain Monte Carlo (MCMC) or other stochastic sampling methods
\cite{FrenkelSmit_MolecularSimulation}. This approach is asymptotically
unbiased, but suffers from the sampling problem: without knowing efficient
moves, MCMC approaches may get stuck in local energy minima for a
long time and fail to converge in practice.

Normalizing flows (NFs) \cite{tabak2010density,tabak2013family,dinh2014nice,rezende2015variational, DinhBengio_RealNVP, PapmakariosEtAl_FlowReview}
combined with importance sampling methods are an alternative approach that
enjoys growing interest in molecular and material sciences and nuclear physics \cite{muller2018neural,LiWang_PRL18_NeuralRenormalizationGroup,noe2019boltzmann,kohler2019equivariant,Albergo_PRD19_FlowLattice,Nicoli_PRE_UnbiasedSampling}.
NFs are learnable invertible functions, usually represented by a neural network, pushing forward a probability density over a latent or ``prior'' space $Z$ towards the target space $X$. Utilizing the change of variable rule these models provide exact densities of generated samples allowing them to be trained by either maximizing the likelihood on data (ML) or minimizing the Kullback-Leibler divergence (KL) towards a target distribution. 

Let $F_{ZX}$ be such a map and its inverse $F_{XZ}=F_{ZX}^{-1}$. We can consider it as composition of $T$ invertible transformation layers $F_{0},...,F_{T}$ with intermediate states $\mathbf{y}_{t}$ given by:
\begin{align}
\mathbf{y}_{t+1}=F_{t}(\mathbf{y}_{t})\:\:\:\:\: & \:\:\:\:\:\mathbf{y}_{t}=F_{t}^{-1}(\mathbf{y}_{t+1})\label{eq:flow_layer_deterministic}
\end{align}

By calling the samples in $Z$ and $X$ also $\mathbf{z}$
and $\mathbf{x}$, respectively, the flow structure is as
follows:
\begin{equation}
\begin{array}{ccccc}
 & F_{0} &  & F_{T-1}\\
\mathbf{z}=\mathbf{y}_{0} & \rightleftarrows & \mathbf{y}_{1}\rightleftarrows\cdots\rightleftarrows\mathbf{y}_{T-1} & \rightleftarrows & \mathbf{y}_{T}=\mathbf{x}\\
 & F_{0}^{-1} &  & F_{T-1}^{-1}
\end{array}
\end{equation}
We suppose each transformation layer is differentiable with a Jacobian
determinant $\left|\det\mathbf{J}_{t}(\mathbf{y})\right|$. This allows to apply the \textit{change of variable }rule:
\begin{equation}
p_{t+1}(\mathbf{y}_{t+1})=p_{t+1}\left(F_{t}(\mathbf{y}_{t})\right)=p_{t}(\mathbf{y}_{t})\left|\det\mathbf{J}_{t}(\mathbf{y}_{t})\right|^{-1}.\label{eq:transformation_random_vars}
\end{equation}
As we often work with log-densities, we abbreviate the log Jacobian
determinant as:
\begin{equation}
\Delta S_{t}=\log\left|\det\mathbf{J}_{t}(\mathbf{y})\right|.
\end{equation}
The log Jacobian determinant of the entire flow is defined by $\Delta S_{ZX}=\sum_{t}\Delta S_{t}(\mathbf{y}_{t})$
and correspondingly $\Delta S_{XZ}$ for the inverse flow.

\paragraph{Unbiased sampling with Boltzmann Generators.} Unbiased sampling is particularly important for applications in physics and chemistry
where unbiased expectation values are required \cite{LiWang_PRL18_NeuralRenormalizationGroup,noe2019boltzmann,Albergo_PRD19_FlowLattice,Nicoli_PRE_UnbiasedSampling}. A Boltzmann generator \cite{noe2019boltzmann} utilizing NFs achieves this by (i) generating one-shot
samples $\mathbf{x}\sim p_{X}(\mathbf{x})$ from the flow and (ii) using a reweighing/resampling procedure respecting weights 
\begin{equation}
w(\mathbf x)=\frac{\mu_X(\mathbf x)}{p_X(\mathbf x)}
\propto\exp\left(-u_{X}(\mathbf{x})+u_{Z}(\mathbf{z})+\Delta S_{ZX}(\mathbf{z})\right), 
\end{equation}
turning these one-shot samples into asymptotically unbiased samples. Reweighing/resampling methods utilized in this context are e.g. \textit{Importance Sampling} \cite{muller2018neural,noe2019boltzmann} or \textit{Neural MCMC} \cite{LiWang_PRL18_NeuralRenormalizationGroup,Albergo_PRD19_FlowLattice,Nicoli_PRE_UnbiasedSampling}.

\paragraph{Training NFs.}\hspace{-0.3cm}
NFs are trained in either ``forward'' or ``reverse'' mode, e.g.:
\begin{enumerate}
\item Density estimation -- given data samples $\mathbf{x}$, train the
flow such that the back-transformed samples $\mathbf{z}=F_{XZ}(\mathbf{x})$
follow a latent distribution $\mu_{Z}(\mathbf{z})$, e.g. $\mu_{Z}(\mathbf{z})=\mathcal{N}(\mathbf{0},\mathbf{I})$. This is done by maximizing the likelihood -- equivalent to minimizing the KL divergence $KL\left[  \mu_X \| p_X \right]$.
\item Sampling of a given target density $\mu_{X}(\mathbf{x})$ -- sample
from the simple distribution $\mu_{Z}(\mathbf{z})$ and minimize a
divergence between the distribution generated by the forward-transformation
$\mathbf{x}=F_{XZ}(\mathbf{z})$ and $\mu_{X}(\mathbf{x})$. A common choice is the reverse KL divergence $KL\left[  p_X \| \mu_X \right]$.
\end{enumerate}
We will use densities interchangeably with energies, defined by the
negative logarithm of the density. The exact prior and target distributions
are:
\begin{align}
\mu_{Z}(\mathbf{z})=Z_{Z}^{-1}\exp(-u_{Z}(\mathbf{z}))\:\:\:\:\: & \:\:\:\:\:\mu_{X}(\mathbf{x})=Z_{X}^{-1}\exp(-u_{X}(\mathbf{x}))\label{eq:exact_distributions}
\end{align}
with generally unknown normalization constants $Z_{Z}$ and $Z_{X}$. As can be shown (Suppl. Material Sec. 1) minimizing $KL\left[  p_X \| \mu_X \right]$ or $KL\left[  \mu_X \| p_X \right]$ corresponds to maximizing the forward or backward weights of samples drawn from $p_X$ or $\mu_X$, respectively.

\paragraph{Topological problems of NFs.}

\begin{figure}
\begin{center}
\includegraphics[width=0.9\columnwidth]{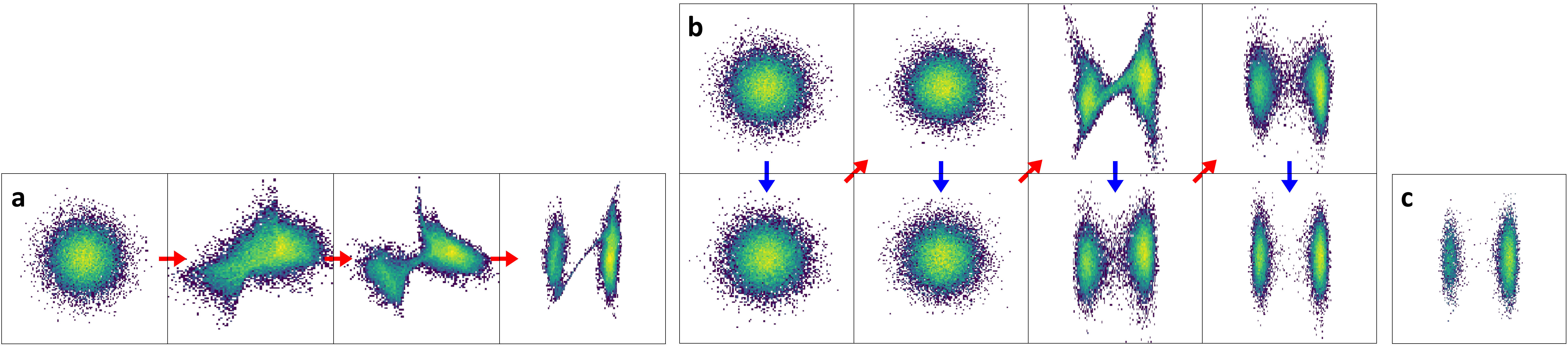}
\end{center}

\caption{\label{fig:double_well_illustration}\textbf{Deterministic versus
stochastic normalizing flow for the double well}. Red arrows indicate
deterministic transformations, blue arrows indicate stochastic dynamics.
\textbf{a}) 3 RealNVP blocks (2 layers each). \textbf{b}) Same with
20 BD steps before or after RealNVP blocks. \textbf{c}) Unbiased sample
from true distribution.}
\end{figure}

A major caveat of sampling with exactly invertible functions for physical problems
are topological constraints. While these can be strong manifold results, e.g., if the sample space is restricted to a non-trivial Lie group \cite{falorsi2018explorations,falorsi2019reparameterizing}, another practical problem are induced Bi-Lipschitz constraints resulting from mapping uni-modal base distributions onto well-separated multi-modal target distributions\cite{cornish2019relaxing}.
For example, when trying to map a unimodal Gaussian distribution to
a bimodal distribution with affine coupling layers, a connection between
the modes remains (Fig. \ref{fig:double_well_illustration}a). 
This representational insufficiency poses serious problems during optimization -- in the bimodal
distribution example, the connection between the density modes seems
largely determined by the initialization and does not move during
optimization, leading to very different results in multiple runs (Suppl. Material, 
Fig. \ref{fig:SI_2well_Reproducibility}).
More powerful coupling layers, e.g., \cite{durkan2019neural}, can mitigate this effect. Yet, as they are still diffeomorphic, strong Bi-Lipschitz requirements can make optimization difficult. This problem can be resolved when relaxing bijectivity of the flow by adding noise as we show in our results. Other proposed solutions are real-and-discrete mixtures of flows \cite{dinh2019rad} or augmentation of the bases space \cite{dupont2019augmented, huang2020augmented} at the cost of losing asymptotically unbiased sampling.

\paragraph{Contributions.}

We show that NFs can be interwoven with stochastic sampling blocks into arbitrary sequences, that together overcome topological constraints and improve expressivity over
deterministic flow architectures (Fig. \ref{fig:double_well_illustration}a, b). Furthermore, NSFs have improved sampling efficiency over pure stochastic sampling as the flow's and sampler's parameters can be  optimized jointly. 

Our main result is that NSFs can be trained in a similar fashion as NFs and exact importance weights for each sample ending in $\mathbf{x}$ can be computed, facilitating asymptotically unbiased sampling from the target density. The approach avoids explicitly computing $p_X(\mathbf{x})$ which would require solving the intractable integral over all stochastic paths ending in $\mathbf{x}$.

We apply the model to the recently introduced problem of asymptotically unbiased sampling of molecular structures with flows \cite{noe2019boltzmann} and show that it significantly improves sampling the multi-modal torsion angle distributions which are the relevant degrees of freedom in the system. We further show the advantage of the method over pure flow-based sampling / MCMC by quantitative comparison on benchmark data sets and on sampling from a VAE's posterior distribution.

\textbf{Code} is available at
\href{https://github.com/noegroup/stochastic_normalizing_flows}{\texttt{github.com/noegroup/stochastic\_normalizing\_flows}}

\section{Stochastic normalizing flows}

\begin{figure}
\centering
\includegraphics[width=0.7\columnwidth]{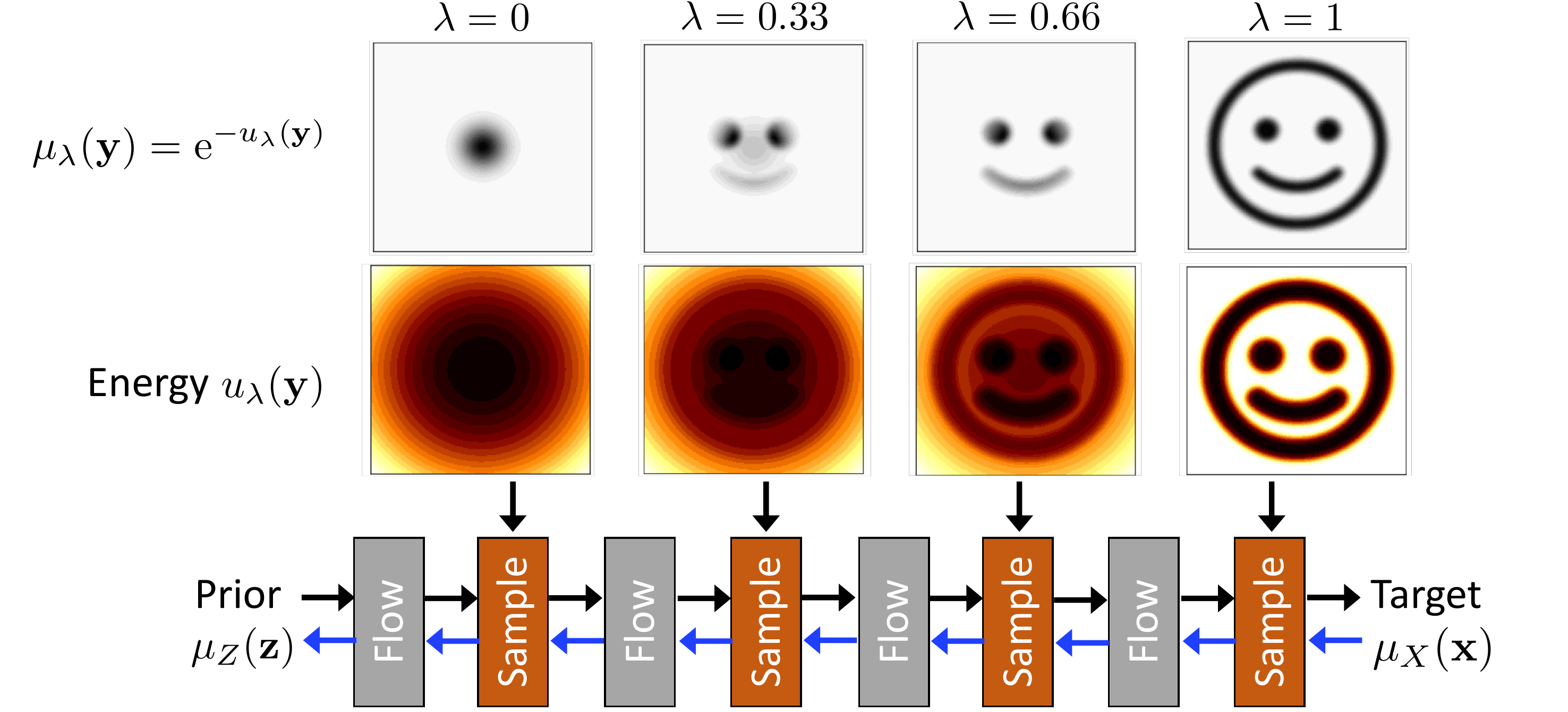}
\caption{
    \label{fig:SNF_scheme}
    \textbf{Schematic for Stochastic Normalizing
    Flow (SNF)}. An SNF transforms a tractable prior $\mu_{Z}(\mathbf{z})\propto\exp(-u_{0}(\mathbf{z}))$
    to a complicated target distribution $\mu_{X}(\mathbf{x})\propto\exp(-u_{1}(\mathbf{x}))$
    by a sequence of deterministic invertible transformations (flows,
    grey boxes) and stochastic dynamics (sample, ochre) that sample with
    respect to a guiding potential $u_{\lambda}(\mathbf{x})$. SNFs can
    be trained and run in forward mode (black) and reverse mode (blue).
}
\end{figure}

A SNF is a sequence of $T$ stochastic and deterministic transformations.
We sample $\mathbf{z}=\mathbf{y}_{0}$ from the prior $\mu_{Z}$,
and generate a forward path $(\mathbf{y}_{1},\ldots,\mathbf{y}_{T})$ resulting in a proposal $\mathbf{y}_{T}$ (Fig. \ref{fig:SNF_scheme}). Correspondingly, latent space samples can be generated by starting
from a sample $\mathbf{x}=\mathbf{y}_{T}$ and invoking the backward
path $(\mathbf{y}_{T-1},\ldots,\mathbf{y}_{0})$. The conditional
forward / backward path probabilities are
\begin{align}
\mathbb{P}_{f}(\mathbf{z}\!=\!\mathbf{y}_{0}\rightarrow\mathbf{y}_{T}\!=\!\mathbf{x})  =\prod_{t=0}^{T-1}q_{t}(\mathbf{y}_{t}\rightarrow\mathbf{y}_{t+1}), 
&
\,\,\,\,\,\,\,\,\mathbb{P}_{b}(\mathbf{x}\!=\!\mathbf{y}_{T}\rightarrow\mathbf{y}_{0}\!=\!\mathbf{z})  =\prod_{t=0}^{T-1}\tilde{q}_{t}(\mathbf{y}_{t+1}\rightarrow\mathbf{y}_{t})
\end{align}
where 
\begin{equation}
    \mathbf{y}_{t+1}|\mathbf{y}_{t}  \sim  q_{t}(\mathbf{y}_{t}\to\mathbf{y}_{t+1})
    \qquad
    \mathbf{y}_{t}|\mathbf{y}_{t+1}  \sim  \tilde{q}_{t}(\mathbf{y}_{t+1}\to\mathbf{y}_{t})
\end{equation}
denote the forward / backward sampling density at step $t$ respectively. If step $t$ is a deterministic transformation $F_{t}$ this simplifies as 
\begin{align*}
\mathbf{y}_{t+1}  \sim \delta\left(\mathbf{y}_{t+1}-F_{t}(\mathbf{y}_{t})\right),\qquad
\mathbf{y}_{t} \sim \delta\left(\mathbf{y}_{t}-F_{t}^{-1}(\mathbf{y}_{t+1})\right).
\end{align*}
In contrast to NFs, the probability that an SNF generates
a sample $\mathbf{x}$ cannot be computed by Eq. (\ref{eq:transformation_random_vars}) but instead involves an integral over all paths that end in $\mathbf{x}$:
\begin{equation}
p_{X}(\mathbf{x})=\int\mu_{Z}(\mathbf{y}_{0})\mathbb{P}_{f}(\mathbf{y}_{0}\rightarrow\mathbf{y}_{T})\:\mathrm{d}\mathbf{y}_{0}\cdots\mathrm{d}\mathbf{y}_{T-1}.\label{eq:generation_probability_direct}
\end{equation}
This integral is generally intractable, thus a feasible training method must avoid using Eq. (\ref{eq:generation_probability_direct}).
Following \cite{NilmeyerEtAl_PNA11_NCMC}, 
we can draw samples $\mathbf{x}\sim\mu_{X}(\mathbf{x})$
by running Metropolis-Hastings moves in the path-space of $(\mathbf{z}=\mathbf{y}_{0},...,\mathbf{y}_{T}=\mathbf{x})$
if we select the backward path probability
$\mu_{X}(\mathbf{x})\mathbb{P}_{b}(\mathbf{x}\rightarrow\mathbf{z})$
as the target distribution and the forward path probability
$\mu_{Z}(\mathbf{z})\mathbb{P}_{f}(\mathbf{z}\rightarrow\mathbf{x})$
as the proposal density.
Since we sample paths independently, it is simpler to assign an unnormalized
importance weight proportional to the acceptance ratio to each sample
path from $\mathbf z=\mathbf y_0$ to $\mathbf x=\mathbf y_T$:
\begin{eqnarray}
w(\mathbf{z}\to\mathbf{x}) 
= \exp\left(-u_{X}(\mathbf{x})+u_{Z}(\mathbf{z})+\sum_{t}\Delta S_{t}(\mathbf{y}_{t})\right)
\propto \frac{\mu_{X}(\mathbf{x})\mathbb{P}_{b}(\mathbf{x}\rightarrow\mathbf{z})}{\mu_{Z}(\mathbf{z})\mathbb{P}_{f}(\mathbf{z}\rightarrow\mathbf{x})}
\label{eq:acceptance_ratio},
\end{eqnarray}
where 
\begin{equation}
\Delta S_{t}=\log\frac{\tilde{q}_{t}(\mathbf{y}_{t+1}\to\mathbf{y}_{t})}{q_{t}(\mathbf{y}_{t}\to\mathbf{y}_{t+1})}\label{eq:SNF_nonequilibrium_work}
\end{equation}
denotes the forward-backward probability ratio of step $t$,
and corresponds to the usual change of variable formula in NF
for deterministic transformation steps (Suppl. Material Sec. 3). 
These weights allow asymptotically unbiased sampling and training of SNFs while avoiding Eq. (\ref{eq:generation_probability_direct}). By changing denominator and numerator in \eqref{eq:acceptance_ratio} we can alternatively obtain the backward weights $w(\mathbf{x}\to\mathbf{z})$.

\paragraph{SNF training.}

As in NFs, the parameters of a SNF can be optimized by minimizing the
Kullback-Leibler divergence between the forward and backward path probabilities, or alternatively maximizing forward and backward path weights as long as we can compute $\Delta S_{t}$ (Suppl. Material Sec 1):
\begin{equation}
J_{\mathrm{KL}}	
=
\mathbb{E}_{\mu_{Z}(\mathbf{z})\mathbb{P}_{f}(\mathbf{z}\to\mathbf{x})}\left[-\log w(\mathbf{z}\to\mathbf{x})\right]
=	\mathrm{KL}\left(\mu_{Z}(\mathbf{z})\mathbb{P}_{f}(\mathbf{z}\to\mathbf{x})||\mu_{X}(\mathbf{x})\mathbb{P}_{b}(\mathbf{x}\to\mathbf{z})\right) + \text{const}.
\end{equation}
In the ideal case of $J_{\mathrm{KL}}=0$, all paths have the same weight $w(\mathbf z\to \mathbf x)=1$
and the independent and identically distributed sampling of $\mu_X$ can be achieved. Accordingly, we can maximize the likelihood of the generating process on data drawn from $\mu_X$ by minimizing:
\begin{equation}
J_{\mathrm{ML}}
=
\mathbb{E}_{\mu_{X}(\mathbf{x})\mathbb{P}_{b}(\mathbf{x}\to\mathbf{z})}\left[-\log w(\mathbf{x}\to\mathbf{z})\right]
=	\mathrm{KL}\left(\mu_{X}(\mathbf{x})\mathbb{P}_{b}(\mathbf{x}\to\mathbf{z})||\mu_{Z}(\mathbf{z})\mathbb{P}_{f}(\mathbf{z}\to\mathbf{x})\right) + \text{const}.
\end{equation}

\paragraph{Variational bound.}
Minimization of the reverse path divergence $J_{KL}$ minimizes an upper bound on the reverse KL divergence between the marginal distributions:
\begin{equation}
\mathrm{KL}\left(p_{X}(\mathbf{x})\parallel\mu_{X}(\mathbf{x})\right)
	\le\mathrm{KL}\left(\mu_{Z}(\mathbf{z})\mathbb{P}_{f}(\mathbf{z}\rightarrow\mathbf{x})\parallel\mu_{X}(\mathbf{x})\mathbb{P}_{b}(\mathbf{x}\rightarrow\mathbf{z})\right)    
\end{equation}
And the same relationship exists between the forward path divergence $J_{ML}$ and the forward KL divergence.
While invoking this variational approximation precludes us from
explicitly computing $p_{X}(\mathbf{x})$ and $\mathrm{KL}\left(p_{X}(\mathbf{x})\parallel\mu_{X}(\mathbf{x})\right)$, we can still generate asymptotically unbiased samples from the target density $\mu_{X}$, unlike in variational inference. 

\paragraph{Asymptotically unbiased sampling.}
As stated in the theorem below (Proof in Suppl. Material. Sec. 2), SNFs are Boltzmann Generators: We can generate asymptotically
unbiased samples of $\mathbf{x}\sim\mu_{X}(\mathbf{x})$ by performing
importance sampling or Neural MCMC using the path weight $w(\mathbf{z}_{k}\rightarrow\mathbf{x}_{k})$
of each path sample $k$.
\begin{theorem}\label{thm:unbiased_sampling}
Let $O$ be a function over $X$.
An asymptotically unbiased estimator
is given by
\begin{eqnarray}
\mathbb{E}_{\mathbf{x} \sim \mu_{X}}\left[ O(\mathbf{x}) \right] & \approx & \frac{\sum_{k}w(\mathbf{z}_{k}\rightarrow\mathbf{x}_{k})\,O(\mathbf{x}_{k})}{\sum_{k}w(\mathbf{z}_{k}\rightarrow\mathbf{x}_{k})},\label{eq:SNF_importance_weight}
\end{eqnarray}
if paths are drawn from the forward path distribution $\mu_{Z}(\mathbf{z})\mathbb{P}_{f}(\mathbf{z}\rightarrow\mathbf{x})$.
\end{theorem}

\section{Implementing SNFs via Annealed Importance Sampling}

\label{sec:Stochastic_Layers}

In this paper we focus on the use of SNFs as samplers of $\mu_{X}(\mathbf{x})$
for problems where the target energy $u_{X}(\mathbf{x})$ is known,
defining the target density up to a constant, and provide an implementation of stochastic blocks via MCMC / LD.
These blocks make local stochastic updates of the
current state $\mathbf{y}$ with respect to some potential $u_{\lambda}(\mathbf{y})$
such that they will asymptotically sample from $\mu_{\lambda}(\mathbf{y})\propto\exp(-u_{\lambda}(\mathbf{y}))$.
While such potentials $u_{\lambda}(\mathbf{y})$ could be learned, a
straightforward strategy is to interpolate between prior
and target potentials
\begin{equation}
u_{\lambda}(\mathbf{y})=(1-\lambda)u_{Z}(\mathbf{y})+\lambda u_{X}(\mathbf{y}),
\end{equation}
similarly as it is done in \textit{annealed importance sampling} \cite{Neal_98_AnnealedImportanceSampling}.
Our implementation for SNFs is thus as follows: deterministic flow layers in-between only have to approximate the partial
density transformation between adjacent $\lambda$ steps while the
stochastic blocks anneal with respect to the given intermediate potential $u_{\lambda}$.
The parameter $\lambda$ could again be learned -- in this paper we simply choose a
linear interpolation along the SNF layers: $\lambda=t/T$.




\paragraph{Langevin dynamics.}

\label{subsec:Stochastic_Layers_Brownian}

Overdamped Langevin dynamics, also known as Brownian dynamics, using an Euler discretization with time step
$\Delta t$, are given by \cite{ErmakYeh_CPL74_BrownianDynamicsIons}:
\begin{align}
\mathbf{y}_{t+1} & =\mathbf{y}_{t}-\epsilon_{t}\nabla u_{\lambda}(\mathbf{y}_{t})+\sqrt{2\epsilon_{t} / \beta}\boldsymbol{\eta}_{t},\label{eq:BD_dynamics1}
\end{align}
where $\boldsymbol{\eta}_{t} \sim \mathcal{N}(0,\mathbf{I})$ is Gaussian noise. In physical
systems, the constant $\epsilon_{t}$ has the form $\epsilon_{t}=\Delta t / \gamma m$
with time step $\Delta t$, friction coefficient $\gamma$ and mass
$m$, and $\beta$ is the inverse temperature (here set to $1$).
The backward step $\mathbf{y}_{t+1}\rightarrow\mathbf{y}_{t}$ is
realized under these dynamics with the backward noise realization
(Suppl. Material Sec. 4 and \cite{NilmeyerEtAl_PNA11_NCMC}):
\begin{equation}
\tilde{\boldsymbol{\eta}}_{t}=\sqrt{\frac{\beta\epsilon_{t}}{2}}\left[\nabla u_{\lambda}(\mathbf{y}_{t})+\nabla u_{\lambda}(\mathbf{y}_{t+1})\right]-\boldsymbol{\eta}_{t}.
\end{equation}
The log path probability ratio is (Suppl. Material Sec. 4):
\begin{equation}
\Delta S_{t}=-\frac{1}{2}\left(\left\Vert \tilde{\boldsymbol{\eta}}_{t}\right\Vert ^{2}-\left\Vert \boldsymbol{\eta}_{t}\right\Vert ^{2}\right).
\end{equation}
We also give the results for non-overdamped Langevin dynamics in Suppl. Material. Sec. 5.

\paragraph{Markov Chain Monte Carlo.}

Consider MCMC methods with a proposal density $q_t$ that satisfies the detailed balance condition
w.r.t. the interpolated density $\mu_\lambda(\mathbf y) \propto \exp(-u_\lambda(\mathbf y))$:
\begin{equation}
\exp(-u_\lambda(\mathbf y_t))q_t(\mathbf y_t\to\mathbf y_{t+1})=
\exp(-u_\lambda(\mathbf y_{t+1}))q_t(\mathbf y_{t+1}\to\mathbf y_t)\label{eq:MCMC_qt}
\end{equation}
We show that for all $q_t$ satisfying (\ref{eq:MCMC_qt}),
including Metropolis-Hastings and Hamiltonian MC moves,
the log path probability ratio is (Suppl. Material Sec. 6 and 7):
\begin{equation}
\Delta S_{t}=u_{\lambda}(\mathbf{y}_{t+1})-u_{\lambda}(\mathbf{y}_{t}),
\end{equation}
if the backward sampling density satisfies $\tilde q_t=q_t$.

\section{Results}


\paragraph{Representational power versus sampling efficiency.}

\begin{figure}
\begin{centering}
\includegraphics[width=1\columnwidth]{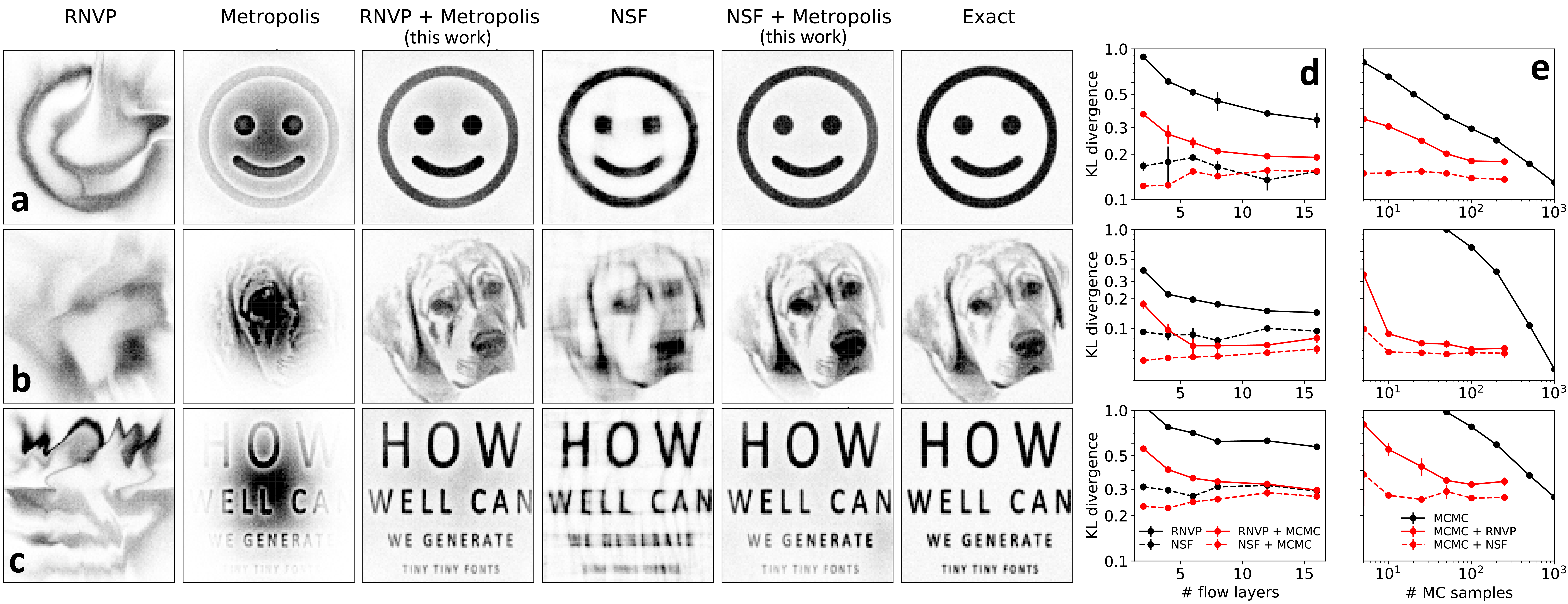}
\end{centering}
\caption{\label{fig:Images}
\textbf{Sampling of two-dimensional densities}. \textbf{a-c)} Sampling of
smiley, dog and text densities with different methods.
Columns: (1) Normalizing Flow with RealNVP layers, (2) Metropolis MC sampling, (3) Stochastic Normalizing Flow combining (1+2), 
(4) neural spline flow (NSF), (5) Stochastic Normalizing Flow combining (1+4),
(6)
Unbiased sample from exact density.
\textbf{d-e)} Compare representative power and statistical
efficiency of different flow methods by showing KL divergence (mean and
standard deviation over 3 training runs) between flow samples and
true density for the three images
from Fig. \ref{fig:Images}. \textbf{d}) Comparison of deterministic
flows (black) and SNF (red) as a function of the number of RealNVP or Neural Spline Flow
transformations. Total number of MC steps in SNF is fixed to 50. \textbf{e})
Comparison of pure Metropolis MC (black) and SNF (red, solid line RealNVP, dashed line Neural spline flow) as a function
of the number of MC steps. Total number of RealNVP or NSF transformations in SNF is fixed to 10.}
\end{figure}

We first illustrate that SNFs can break topological constraints and
improve the representational power of deterministic normalizing flows at a given network size
and at the same time beat direct MCMC in terms of sampling efficiency.
To this end we use images to define complex two-dimensional densities
(Fig. \ref{fig:Images}a-c, ``Exact'') as target densities $\mu_{X}(\mathbf{x})$
to be sampled. Note that a benchmark aiming at generating high-quality
images would instead represent the image as a high-dimensional pixel
array. We compare three types of flows with 5 blocks each trained by samples
from the exact density (details in Suppl. Material Sec. 9): 

\begin{enumerate}
    \item Normalizing flow with 2 swapped coupling layers (RealNVP or neural spline flow) per block
    \item Non-trainable stochastic flow with 10 Metropolis MC steps per block
    \item SNF with both, 2 swapped coupling layers and 10 Metropolis MC steps per block.
\end{enumerate}

The pure Metropolis MC flow suffers from sampling problems -- density
is still concentrated in the image center from the prior. Many more
MC steps would be needed to converge to the exact density (see below).
The RealNVP normalizing flow architecture \cite{DinhBengio_RealNVP} has limited
representational power, resulting in a ``smeared out'' image that
does not resolve detailed structures (Fig. \ref{fig:Images}a-c, RNVP).
As expected, neural spline flows perform significantly better on the 2D-images than RealNVP flows, but at the chosen network architecture their ability to resolve fine details and round shapes is still limite (See dog and small text in Fig. \ref{fig:Images}c, NSF).
Note that the representational power for all flow architectures tend to increase
with depth - here we compare the performance of different architectures at fixed depth
and similar computational cost.

In contrast, SNFs achieve high-quality approximations although they simply combine
the same deterministic and stochastic flow components that fail individually
in the SNF learning framework 
(Fig. \ref{fig:Images}a-c, RNVP+Metropolis and NSF+Metropolis).
This indicates that the SNF succeeds in performing
the large-scale probability mass transport with the trainable flow
layers and sampling the details with Metropolis MC.

Fig. \ref{fig:Images}d-e quantifies these impressions by computing
the KL divergence between generated densities $p_{X}(\mathbf{x})$
and exact densities $\mu_{X}(\mathbf{x})$. Both normalizing flows
and SNFs improve with greater depth, but SNFs achieve significantly
lower KL divergence at a fixed network depth (Fig. \ref{fig:Images}d). 
Note that both RealNVP
and NSFs improve significantly when stochasticty is added.

Moreover, SNFs have higher statistical efficiency than pure Metropolis MC flows. 
Depending on the example and flow architecture, 1-2 orders of magnitude 
more Metropolis MC steps are needed to achieve similar KL divergence as with an SNF. 
This demonstrates that the large-scale probability transport learned by the trainable
deterministic flow blocks in SNFs significantly helps with the sampling.

Importantly, adding stochasticity is very inexpensive. Although
every MCMC or Langevin integration step adds a neural network layer, these layers are
very lightweighted, and have only linear computational complexity in the number of 
dimensions. As an example, for our SNF implementation of the examples in Fig. \ref{fig:Images} we can add 10-20 stochastic layers to each trainable normalizing flow layer before the computational cost increases by a factor of 2 (Suppl. Material Fig. \ref{fig:Images_timings}).

\paragraph{SNFs as asymptotically unbiased samplers.}

\begin{figure}[t]
\centering
\noindent \includegraphics[width=1.0\columnwidth]{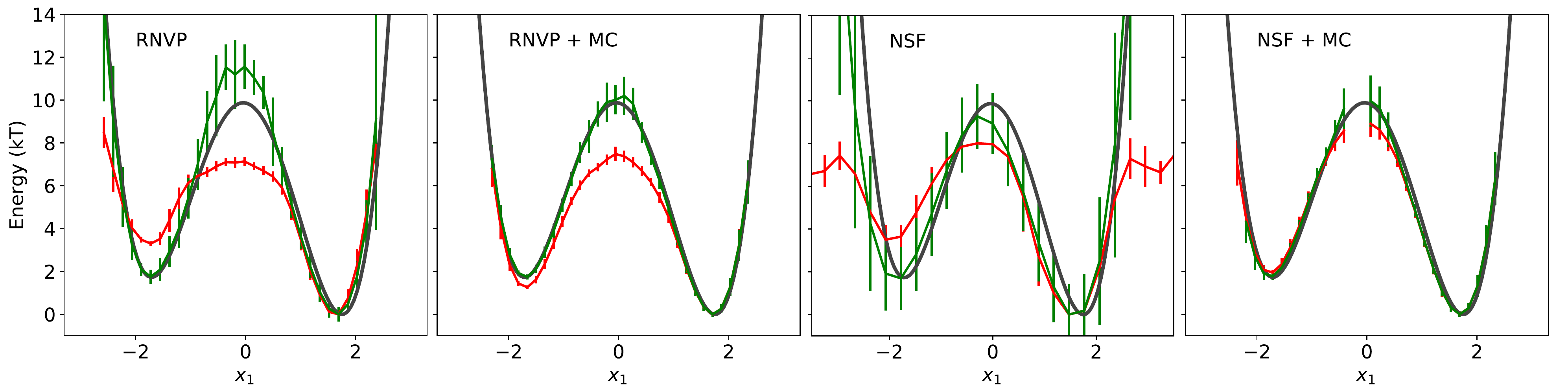}

\caption{
\label{fig:reweighting_2well}\textbf{Reweighting results for the
double well potential} (see also Fig. \ref{fig:double_well_illustration}).
Free energy along $x_{1}$ (negative log of marginal density)
for deterministic normalizing flows (RNVP, NSF) and SNFs (RNVP+MC, NSF+MC). Black:
exact energy, red: energy of proposal density $p_{X}(\mathbf{x})$,
green: reweighted energy using importance sampling. }
\end{figure}

We demonstrate that SNFs can be used as Boltzmann Generators, i.e., to sample target densities without
asymptotic bias by revisiting the double-well example (Fig. \ref{fig:double_well_illustration}).
Fig. \ref{fig:reweighting_2well} (black) shows the free energies
(negative marginal density) along the double-well coordinate $x_{1}$.
Flows with 3 coupling layer blocks (RealNVP or neural spline flow) are trained summing forward and reverse KL divergence as a joint loss
using either data from a biased distribution, or with the unbiased
distribution (Details in Suppl. Material Sec. 9). 
Due to limitations in representational power the generation probability $p_{X}(\mathbf{x})$ will be biased -- even when explicitly minimizing the KL divergence w.r.t. the true unbiased distribution in the joint loss.
By relying on importance sampling we can turn the flows into Boltzmann Generators \cite{noe2019boltzmann} in order to obtain unbiased estimates.
Indeed all generator
densities $p_{X}(\mathbf{x})$ can be reweighted to an estimate of
the unbiased density $\mu_{X}(\mathbf{x})$ whose free energies are
within statistical error of the exact result (Fig. \ref{fig:reweighting_2well},
red and green).

We inspect the bias, i.e. the error of the mean estimator, and the statistical uncertainty ($\sqrt{\mathrm{var}}$) of the free energy in $x_1 \in \{-2.5, 2.5\}$ with and without reweighting using a fixed number of samples (100,000).
Using SNFs with Metropolis MC steps, both biases and uncertainties are reduced by half
compared to purely deterministic flows (Table \ref{tab:reweighting_2well}). 
Note that neural spline flows perform better than RealNVP without reweighting, but significantly worse with reweighting - presumably because the sharper features representable by splines can be detrimental for reweighting weights. With stochastic layers, both RealNVP and neural spline flows perform approximately equally well.

The differences between multiple runs (see standard deviations of the uncertainty
estimate) also reduce significantly, i.e. SNF results are more reproducible
than RealNVP flows, confirming that the training problems caused by
the density connection between both modes (Fig. \ref{fig:double_well_illustration},
Suppl. Material Fig. \ref{fig:SI_2well_Reproducibility}) can be reduced.
Moreover, the sampling performance of SNF can be further improved by optimizing MC step sizes based on loss functions $J_{KL}$ and $J_{ML}$ (Suppl. Material Table \ref{tab:adaptive_mc}).

Reweighting reduces the bias at the expense of a higher variance. Especially in physics applications, a small or asymptotically zero bias is often very important, and the variance can be reduced by generating more samples from the trained flow, which is relatively cheap and parallel.

\begin{table}[h]
\caption{\textbf{Unbiased sampling for double well potential:} mean
uncertainty of the reweighted energy along $x_{1}$ averaged over
10 independent runs ($\pm$ standard deviation).}
\label{tab:reweighting_2well}
    \centering
    \begin{tabular}{l|ccc|ccc}
     \toprule 
  & \multicolumn{3}{|c}{not reweighted} & \multicolumn{3}{|c}{reweighted}\\ 
  & bias & $\sqrt{\textrm{var}}$ & $\sqrt{\textrm{bias}^2 \hspace{-0.1cm}+\hspace{-0.1cm} \textrm{var}}$ & bias & $\sqrt{\textrm{var}}$ & $\sqrt{\textrm{bias}^2 \hspace{-0.1cm}+\hspace{-0.1cm} \textrm{var}}$\\ 
\midrule
RNVP    & $1.4\pm0.6$ & $0.4\pm0.1$ & $1.5\pm0.5$ & $0.3\pm0.2$ & $1.1\pm0.4$ & $1.2\pm0.4$\\ 
\textbf{RNVP + MC} & $1.5\pm0.2$  & $0.3\pm0.1$ & $1.5\pm0.2$ & $\bf0.2\pm0.1$ & $\bf0.6\pm0.1$ & $\bf0.6\pm0.1$\\
NSF     & $0.8\pm0.4$ & $1.0\pm0.2$ & $1.3\pm0.3$ & $0.6\pm0.2$ & $2.1\pm0.4$ & $2.2\pm0.5$ \\ 
\textbf{NSF + MC}  & $0.4\pm0.3$ & $0.5\pm0.1$ & $0.7\pm0.2$ & $\bf 0.1\pm0.1$ & $\bf 0.6\pm0.2$ & $\bf 0.6\pm0.2$ \\

\bottomrule
\end{tabular}
\end{table}

\paragraph{Alanine dipeptide.}

\begin{figure}[h]
\centering
\textbf{a}\includegraphics[clip,width=0.35\columnwidth]{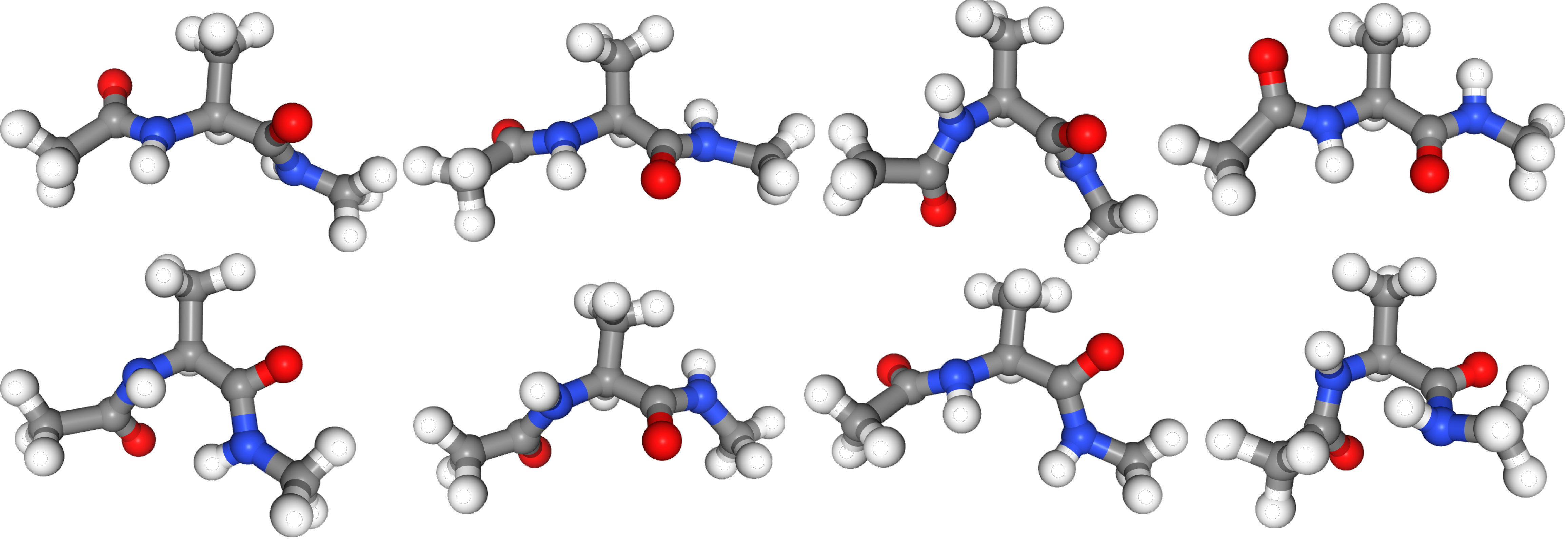}
 \textbf{b}\includegraphics[viewport=100 0 1000 200, clip, width=0.60\columnwidth]{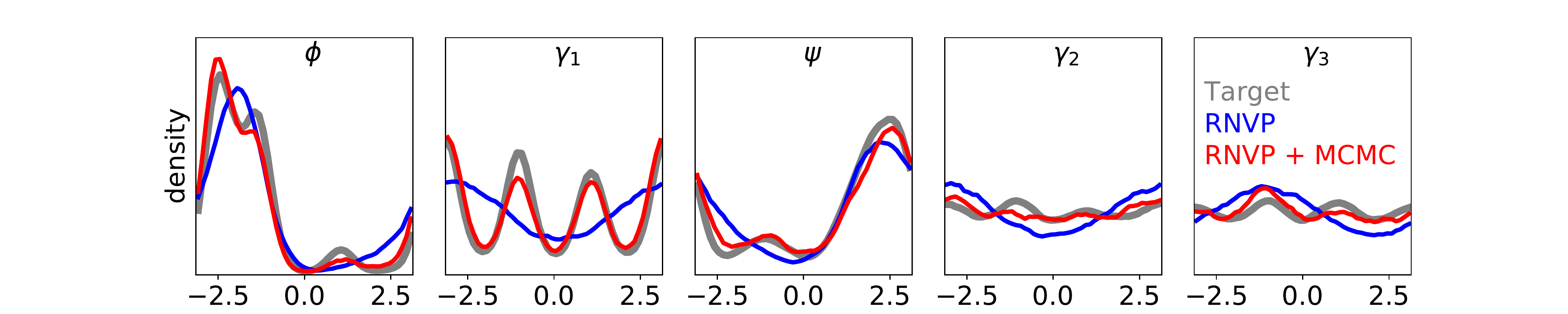}

\caption{\label{fig:Ala2}\textbf{Alanine dipeptide} sampled with deterministic
normalizing flows and stochastic normalizing flows. \textbf{a}) One-shot
SNF samples of alanine dipeptide structures. \textbf{b}) Energy (negative
logarithm) of marginal densities in 5 unimodal torsion angles (top)
and all 5 multimodal torsion angles (bottom).}
\end{figure}

We further evaluate SNFs on density
estimation and sampling of molecular structures from a simulation
of the alanine dipeptide molecule in vacuum (Fig. \ref{fig:Ala2}).
The molecule has 66 dimensions in $\mathbf{x}$, and we augment it
with 66 auxiliary dimensions in a second channel $\mathbf{v}$, similar
to ``velocities'' in a Hamiltonian flow framework \cite{TothEtAl_HamiltonianGenerativeNetworks},
resulting in 132 dimensions total. The target density is given by
$\mu_{X}(\mathbf{x},\mathbf{v})=\exp\left(-u(\mathbf{x})-\frac{1}{2}\left\Vert \mathbf{v}\right\Vert ^{2}\right)$,
where $u(\mathbf{x})$ is the potential energy of the molecule and
$\frac{1}{2}\left\Vert \mathbf{v}\right\Vert ^{2}$ is the kinetic
energy term. $\mu_{Z}$ is an isotropic Gaussian normal distribution
in all dimensions. We utilize the invertible coordinate transformation
layer introduced in \cite{noe2019boltzmann} in order to transform
$\mathbf{x}$ into normalized bond, angle and torsion coordinates.
RealNVP transformations act between the $\mathbf{x}$ and $\mathbf{v}$
variable groups Details in Suppl. Material Sec. 9).

We compare deterministic normalizing flows using 5 blocks of 2 RealNVP
layers with SNFs that additionally use 20 Metropolis MC steps in each
block totalling up to 100 MCMC steps in one forward pass. Fig. \ref{fig:Ala2}a shows random structures sampled by the
trained SNF. Fig. \ref{fig:Ala2}b shows marginal densities in all
five multimodal torsion angles (backbone angles $\phi$, $\psi$ and
methyl rotation angles $\gamma_{1}$, $\gamma_{2}$, $\gamma_{3}$).
While the RealNVP networks that are state of the art for this problem
miss many of the modes, the SNF resolves the multimodal structure
and approximates the target distribution better, as quantified in
the KL divergence between the generated and target marginal distributions
(Table \ref{tab:Ala2}). 

\begin{table}[h]
\caption{\textbf{Alanine dipeptide}: KL-divergences
of RNVP flow and SNF (RNVP+MCMC) between generated and target distributions
for all multimodal torsion angles. Mean and standard deviation from
3 independent runs.}
\label{tab:Ala2}
    \centering
    
    \begin{tabular}{cccccc}
    \toprule
    { KL-div.} & { $\phi$} & { $\gamma_{1}$} & { $\psi$} & { $\gamma_{2}$} & { $\gamma_{3}$}\\
    \midrule
    { RNVP} & { 1.69\textpm 0.03} & { 3.82\textpm 0.01} & { 0.98\textpm 0.03} & { 0.79\textpm 0.03} & { 0.79\textpm 0.09} \\
    { SNF} & \textbf{ 0.36}{ \textpm }\textbf{ 0.05} & \textbf{ 0.21}{ \textpm 0.01} & \textbf{ 0.27}{ \textpm 0.03} & \textbf{ 0.12}{ \textpm 0.02} & \textbf{ 0.15}{ \textpm 0.04}\\
    \bottomrule
    \end{tabular}
\end{table}

\paragraph{Variational Inference.}

Finally, we use normalizing flows to model the latent space distribution
of a variational autoencoder (VAE) 
, as suggested in \cite{rezende2015variational}. Table \ref{tab:VAE}
shows results for the variational bound and the log likelihood on
the test set for \href{http://yann.lecun.com/exdb/mnist/}{MNIST}
\cite{LeCun_ProcIEEE89_ConvNets} and \href{https://github.com/zalandoresearch/fashion-mnist}{Fashion-MNIST}
\cite{Xiao_17_FashionMNIST}.
For a 50-dimensional latent space we compare a six-layer RNVP to MCMC using overdamped Langevin dynamics as proposal (MCMC) and a SNF combining both (RNVP+MCMC). 
Both sampling and the deterministic flow improve over a naive VAE using a reparameterized diagonal Gaussian variational posterior distribution, while the SNF outperforms both, RNVP and MCMC. See Suppl. Material Sec. 8 for details.


\begin{table}
\caption{\label{tab:VAE}\textbf{Variational inference using VAEs with stochastic normalizing flows}: $J_{KL}$: variational bound of the KL-divergence
computed during training. NLL: negative log likelihood of test set.
}
\centering
\begin{tabular}{crrrr}
\toprule
\multicolumn{1}{c}{} & \multicolumn{2}{c}{{  MNIST}} & \multicolumn{2}{c}{{  Fashion-MNIST}}\\
\midrule
\multicolumn{1}{c}{} & {  $J_{KL}$} & {  NLL} & {  $J_{KL}$} & {  NLL}\\
\midrule
{  Naive (Gaussian)} & {  108.4\textpm 24.3} & {  98.1\textpm 4.2} & {  241.3\textpm 7.4} & {  238.0\textpm 2.9}\\
{  RNVP} & {  91.8\textpm 0.4} & {  87.0\textpm 0.2} & {  233.7\textpm 0.1} & {  231.4\textpm 0.2}\\
{  MCMC} & {  102.1\textpm 8.0} & {  96.2\textpm 1.9} & {  234.7\textpm 0.4} & {  235.2\textpm 2.4}\\
{  SNF} & \textbf{  89.7\textpm 0.1} & \textbf{  86.8\textpm 0.1} & \textbf{  232.4\textpm 0.2} & \textbf{  230.9\textpm 0.2}\\
\bottomrule
\end{tabular}
\end{table}

\section{Related work}

The cornerstone of our work is nonequilibrium statistical mechanics.
Particularly important is Nonequilibrium Candidate Monte Carlo (NCMC)
\cite{NilmeyerEtAl_PNA11_NCMC}, which provides the theoretical framework
to compute SNF path likelihood ratios. However, NCMC is for fixed
deterministic and stochastic protocols, while we generalize this into a generative model by substituting fixed protocols with trainable layers and deriving an unbiased optimization procedure. 

Neural stochastic differential equations learn optimal parameters of designed stochastic processes
from observations along the path \cite{TzenRaginsky_NeuralSDE,JiaBenson_NeuralJumpSDE,LiuEtAl_NeuralSDE,LiEtAl_ScalableGradientSDE},
but are not designed for marginal density estimation or asymptotically
unbiased sampling. It has been demonstrated that combining learnable
proposals/transformations with stochastic sampling techniques can
improve expressiveness of the proposals \cite{salimans2015markov,levy2017generalizing,song2017nice,hoffman2019neutra,Hoffman_ICML17_LatentGaussianModels}.
Yet, these contributions do not provide an exact reweighing scheme
based on a tractable model likelihood and do not provide efficient algorithms to optimize arbitrary sequences of transformation or sampling steps end-to-end efficiently. 
These methods can be seen as instances of SNFs with specific choice of deterministic transformations and / or stochastic blocks and model-specific optimizations - see Suppl. Material Table \ref{tab:method_comparison}) for a categorization. 
While our experiments focus on nontrainable stochastic blocks,
the proposal densities of MC steps can also be optimized within the framework of SNFs
as shown in Suppl. Material Table \ref{tab:adaptive_mc}.

An important aspect of SNFs compared to trainable Monte-Carlo kernels such as A-NICE-MC \cite{song2017nice} is the use of detailed balance (DB). While Monte-Carlo frameworks are usually designed to use DB in each step, SNFs rely on path-based detailed balance between the prior and the target density. This means that SNFs can also perform nonequilibrium moves along the transformation, as done by Langevin dynamics without acceptance step and by the deterministic flow transformations such as RealNVP and neural spline flows.

More closely related is \cite{sohl2015deep} which uses of stochastic
flows for density estimation and trains diffusion kernels by maximizing
a variational bound of the model likelihood. Their derivation using
stochastic paths is similar to ours and this work can be seen as a
special instance of SNFs, but it does not consider more general stochastic
and deterministic building blocks and does not discuss the problem
of asymptotically unbiased sampling of a target density. Ref. \cite{chen2017continuous}
proposes a learnable stochastic process by integrating Langevin dynamics
with learnable drift and diffusion term. This approach is in
a spirit similar as our proposed method, but requires variational approximation
of the generative distribution and it
has not been worked out how it could be used as a building block within
a NF. The approach of \cite{gu2019dynamical} combines NF layers with Langevin dynamics, yet approximates the intractable integral with MC samples which we can avoid utilizing the path-weight derivation. Finally, \cite{hodgkinson2020stochastic} propose a stochastic extension to neural ODEs \cite{chen2018neural} which can then be trained as samplers. This approach to sampling is very general yet requires costly integration of a SDE which we can avoid by combining simple NFs with stochastic layers.

\section{Conclusions}

We have introduced stochastic normalizing flows (SNFs) that combine
both stochastic processes and invertible deterministic transformations
into a single learning framework. 
By leveraging nonequilibrium statistical mechanics we show that SNFs can efficiently be trained to sample asymptotically unbiased from target densities. This can be done by utilizing path probability ratios and avoiding intractabe marginalization.
Besides possible applicability in classical machine learning domains such as variational and Bayesian inference, we believe that the latter property can make SNFs a key component in the efficient sampling of many-body physics systems.
In future research we aim to apply SNFs with many stochastic sampling steps to accurate  large-scale sampling of molecules.

\newpage

\section*{Broader Impact}
The sampling of probability distributions defined by energy models is a
key step in the rational design of pharmacological drug molecules for
disease treatment, and the design of new materials, e.g., for energy
storage. Currently such sampling is mostly done by Molecular Dynamics (MD)
and MCMC simulations, which is in many cases limited by computational
resources and generates extremely high energy costs. For example, the direct simulation of a single protein-drug binding and dissociation event could
require the computational time of an entire supercomputer for a year.
Developing machine learning (ML) approaches to solve this problem more
efficiently and go beyond existing enhanced sampling methods is therefore of importance for applications in
medicine and material science and has potentially far-reaching
societal consequences for developing better treatments and reducing
energy consumption. 
Boltzmann Generators, i.e. the combination of Normalizing Flows (FNs) and
resampling/reweighting are a new and promising ML approach to this problem
and the current paper adds a key technology to overcome some of the
previous limitations of NFs for this task.

A risk of the method is that flow-based sampling bears the risk
that non-ergodic samplers can be constructed, i.e. samplers that are not
guaranteed to sample from the target distribution even in the limit of
long simulation time. From such an incomplete sample, wrong conclusions
can be drawn. While incomplete sampling is also an issue with MD/MCMC,
it is well understood how to at least ensure ergodicity of these methods
in the asymptotic limit, i.e. in the limit of generating enough data.
Further research is needed to obtain similar results with normalizing flows.

\paragraph{Acknowledgements.}
Special thanks to José Migual Hernández Lobarto (University of Cambridge) for his valuable input on the path probability ratio for MCMC and HMC. We acknowledge funding from the European Commission (ERC CoG 772230 ScaleCell), Deutsche Forschungsgemeinschaft (GRK DAEDALUS, SFB1114/A04), the Berlin Mathematics center MATH+ (Project AA1-6 and EF1-2) and the Fundamental Research Funds for the Central Universities of China (22120200276).

\printbibliography

@article{Xiao_17_FashionMNIST,
	Author = {Han Xiao and Kashif Rasul and Roland Vollgraf},
	Journal = {arXiv preprint arXiv:cs.LG/1708.07747},
	Title = {Fashion-MNIST: a Novel Image Dataset for Benchmarking Machine Learning Algorithms},
	Year = {2017}}

@article{PapmakariosEtAl_FlowReview,
	Author = {George Papamakarios and Eric Nalisnick and Danilo Jimenez Rezende and Shakir Mohamed and Balaji Lakshminarayanan},
	Journal = {arXiv preprint arXiv:1912.02762},
	Title = {Normalizing Flows for Probabilistic Modeling and Inference},
	Year = {2019}}

@article{Nicoli_PRE_UnbiasedSampling,
	Author = {Kim A. Nicoli and Shinichi Nakajima and Nils Strodthoff and Wojciech Samek and Klaus-Robert M{\"u}ller and Pan Kessel},
	Journal = {arXiv:1910.13496},
	Title = {Asymptotically unbiased estimation of physical observables with neural samplers},
	Year = {2019}}

@article{LiWang_PRL18_NeuralRenormalizationGroup,
	Author = {Shuo-Hui Li and Lei Wang},
	Journal = {Phys. Rev. Lett. },
	Pages = {260601},
	Title = {Neural Network Renormalization Group},
	Volume = {121},
	Year = {2018}}

@article{Albergo_PRD19_FlowLattice,
	Author = {M. S. Albergo and G. Kanwar and P. E. Shanahan},
	Journal = {Phys. Rev. D},
	Pages = {034515},
	Title = {Flow-based generative models for Markov chain Monte Carlo in lattice field theory},
	Volume = {100},
	Year = {2019}}

@article{KingmaBa_ADAM,
	Author = {D. P. Kingma and J. Ba},
	Journal = {arXiv:1412.6980},
	Title = {Adam: A Method for Stochastic Optimization},
	Year = {2014}}

@article{SongEtAl_NIPS17_ANiceMC,
	Author = {J. Song and S. Zhao and S. Ermon},
	Journal = {NIPS},
	Title = {A-NICE-MC: Adversarial Training for MCMC},
	Year = {2017}}

@article{DinhBengio_RealNVP,
	Author = {Laurent Dinh and Jasha Sohl-Dickstein and Samy Bengio},
	Journal = {arXiv:1605.08803},
	Title = {Density estimation using Real NVP},
	Year = {2016}}

@inproceedings{KingmaWelling_ICLR14_VAE,
	Author = {D. P. Kingma and M. Welling},
	Booktitle = {Proceedings of the 2nd International Conference on Learning Representations (ICLR), arXiv:1312.6114},
	Title = {Auto-Encoding Variational Bayes},
	Year = {2014}}

@article{LeCun_ProcIEEE89_ConvNets,
	Author = {Y. LeCun and L. Bottou and Y. Bengio and P. Haffner},
	Journal = {Proc. IEEE},
	Pages = {2278-2324},
	Title = {Gradient-based learning applied to document recognition},
	Volume = {86},
	Year = {1998}}

@article{ErmakYeh_CPL74_BrownianDynamicsIons,
	Author = {D. L. Ermak and Y. Yeh},
	Journal = {Chem. Phys. Lett.},
	Pages = {243-248},
	Title = {Equilibrium electrostatic effects on the behavior of polyions in solution: polyion-mobile ion interaction},
	Volume = {24},
	Year = {1974}}

@article{NilmeyerEtAl_PNA11_NCMC,
	Author = {Jerome P. Nilmeier and Gavin E. Crooks and David D. L. Minh and John D. Chodera},
	Journal = {Proc. Natl. Acad. Sci. USA},
	Pages = {E1009-E1018},
	Title = {{Nonequilibrium candidate Monte Carlo is an efficient tool for equilibrium simulation}},
	Volume = {108},
	Year = {2011}}

@book{FrenkelSmit_MolecularSimulation,
	Author = {Daan Frenkel and Brerend Smit},
	Publisher = {Academic Press},
	Title = {Understanding molecular simulation},
	Year = {2001}}

@article{amber,
	Author = {Pearlman, D. A. and Case, D. A. and Caldwell, J. W. and Ross, W. R. and Iii, Cheatham T. E. and Debolt, S. and Ferguson, D. and Seibel, G. and Kollman, P.},
	Journal = {Comp. Phys. Commun.},
	Pages = {1--41},
	Title = {{AMBER, a computer program for applying molecular mechanics, normal mode analysis, molecular dynamics and free energy calculations to elucidate the structures and energies of molecules}},
	Volume = {91},
	Year = {1995}}

@article{gu2019dynamical,
  title={Dynamical Sampling with Langevin Normalization Flows},
  author={Gu, Minghao and Sun, Shiliang and Liu, Yan},
  journal={Entropy},
  volume={21},
  number={11},
  pages={1096},
  year={2019},
  publisher={Multidisciplinary Digital Publishing Institute}
}

@article{hodgkinson2020stochastic,
  title={Stochastic Normalizing Flows},
  author={Hodgkinson, Liam and van der Heide, Chris and Roosta, Fred and Mahoney, Michael W},
  journal={arXiv preprint arXiv:2002.09547},
  year={2020}
}

@article{cornish2019relaxing,
  title={Relaxing bijectivity constraints with continuously indexed normalising flows},
  author={Cornish, Rob and Caterini, Anthony L and Deligiannidis, George and Doucet, Arnaud},
  journal={arXiv preprint arXiv:1909.13833},
  year={2019}
}

@article{huang2020augmented,
  title={Augmented normalizing flows: Bridging the gap between generative flows and latent variable models},
  author={Huang, Chin-Wei and Dinh, Laurent and Courville, Aaron},
  journal={arXiv preprint arXiv:2002.07101},
  year={2020}
}

@article{Hoffman_ICML17_LatentGaussianModels,
	Author = {Matthew D. Hoffman},
	Journal = {International Conference on Machine Learning},
	Pages = {1510-1519},
	Title = {Learning deep latent Gaussian models with Markov chain Monte Carlo},
	Year = {2017}}

@article{Neal_98_AnnealedImportanceSampling,
	Author = {Radford M. Neal},
	Journal = {arXiv preprint arXiv:physics/9803008},
	Title = {Annealed Importance Sampling},
	Year = {1998}}

@article{TothEtAl_HamiltonianGenerativeNetworks,
	Author = {Peter Toth and Danilo Jimenez Rezende and Andrew Jaegle and S{\'e}bastien Racani{\`e}re and Aleksandar Botev and Irina Higgins},
	Journal = {arXiv preprint arXiv:1909.13789},
	Title = {Hamiltonian Generative Networks},
	Year = {2019}}

@article{TzenRaginsky_NeuralSDE,
	Author = {Belinda Tzen and Maxim Raginsky},
	Journal = {arXiv preprint arXiv:1905.09883},
	Title = {Neural stochastic differential equations: Deep latent Gausian models in the diffusion limit},
	Year = {2019}}

@article{LiuEtAl_NeuralSDE,
	Author = {Xuanqing Liu and Si Si and Qin Cao and Sanjiv Kumar and Cho-Jui Hsieh},
	Journal = {arXiv preprint arXiv:1906.02355},
	Title = {Neural SDE: Stabilizing neural ode networks with stochastic noise},
	Year = {2019}}

@article{JiaBenson_NeuralJumpSDE,
	Author = {Junteng Jia and Austin R. Benson},
	Journal = {arXiv preprint arXiv:1905.10403},
	Title = {Neural Jump Stochastic Differential Equations},
	Year = {2019}}

@article{sohl2015deep,
	Author = {Sohl-Dickstein, Jascha and Weiss, Eric A and Maheswaranathan, Niru and Ganguli, Surya},
	Journal = {arXiv preprint arXiv:1503.03585},
	Title = {Deep unsupervised learning using nonequilibrium thermodynamics},
	Year = {2015}}

@article{levy2017generalizing,
	Author = {Levy, Daniel and Hoffman, Matthew D and Sohl-Dickstein, Jascha},
	Journal = {arXiv preprint arXiv:1711.09268},
	Title = {Generalizing hamiltonian monte carlo with neural networks},
	Year = {2017}}

@inproceedings{song2017nice,
	Author = {Song, Jiaming and Zhao, Shengjia and Ermon, Stefano},
	Booktitle = {Advances in Neural Information Processing Systems},
	Pages = {5140--5150},
	Title = {A-nice-mc: Adversarial training for mcmc},
	Year = {2017}}

@article{chen2017continuous,
	Author = {Chen, Changyou and Li, Chunyuan and Chen, Liqun and Wang, Wenlin and Pu, Yunchen and Carin, Lawrence},
	Journal = {arXiv preprint arXiv:1709.01179},
	Title = {Continuous-time flows for efficient inference and density estimation},
	Year = {2017}}

@inproceedings{salimans2015markov,
	Author = {Salimans, Tim and Kingma, Diederik and Welling, Max},
	Booktitle = {International Conference on Machine Learning},
	Pages = {1218--1226},
	Title = {Markov chain monte carlo and variational inference: Bridging the gap},
	Year = {2015}}

@article{hoffman2019neutra,
	Author = {Hoffman, Matthew and Sountsov, Pavel and Dillon, Joshua V and Langmore, Ian and Tran, Dustin and Vasudevan, Srinivas},
	Journal = {arXiv preprint arXiv:1903.03704},
	Title = {NeuTra-lizing bad geometry in Hamiltonian Monte Carlo using neural transport},
	Year = {2019}}

@article{dinh2019rad,
	Author = {Dinh, Laurent and Sohl-Dickstein, Jascha and Pascanu, Razvan and Larochelle, Hugo},
	Journal = {arXiv preprint arXiv:1903.07714},
	Title = {A RAD approach to deep mixture models},
	Year = {2019}}

@inproceedings{dupont2019augmented,
	Author = {Dupont, Emilien and Doucet, Arnaud and Teh, Yee Whye},
	Booktitle = {Advances in Neural Information Processing Systems},
	Pages = {3134--3144},
	Title = {Augmented neural odes},
	Year = {2019}}

@article{falorsi2018explorations,
	Author = {Falorsi, Luca and de Haan, Pim and Davidson, Tim R and De Cao, Nicola and Weiler, Maurice and Forr{\'e}, Patrick and Cohen, Taco S},
	Journal = {arXiv preprint arXiv:1807.04689},
	Title = {Explorations in homeomorphic variational auto-encoding},
	Year = {2018}}

@article{falorsi2019reparameterizing,
	Author = {Falorsi, Luca and de Haan, Pim and Davidson, Tim R and Forr{\'e}, Patrick},
	Journal = {arXiv preprint arXiv:1903.02958},
	Title = {Reparameterizing Distributions on Lie Groups},
	Year = {2019}}

@article{tabak2010density,
	Author = {Tabak, Esteban G and Vanden-Eijnden, Eric and others},
	Journal = {Communications in Mathematical Sciences},
	Number = {1},
	Pages = {217--233},
	Publisher = {International Press of Boston},
	Title = {Density estimation by dual ascent of the log-likelihood},
	Volume = {8},
	Year = {2010}}

@article{tabak2013family,
	Author = {Tabak, Esteban G and Turner, Cristina V},
	Journal = {Communications on Pure and Applied Mathematics},
	Number = {2},
	Pages = {145--164},
	Publisher = {Wiley Online Library},
	Title = {A family of nonparametric density estimation algorithms},
	Volume = {66},
	Year = {2013}}

@article{dinh2014nice,
	Author = {Dinh, Laurent and Krueger, David and Bengio, Yoshua},
	Journal = {arXiv preprint arXiv:1410.8516},
	Title = {Nice: Non-linear independent components estimation},
	Year = {2014}}

@article{muller2018neural,
	Author = {M{\"u}ller, Thomas and McWilliams, Brian and Rousselle, Fabrice and Gross, Markus and Nov{\'a}k, Jan},
	Journal = {arXiv preprint arXiv:1808.03856},
	Title = {Neural importance sampling},
	Year = {2018}}

@inproceedings{durkan2019neural,
	Author = {Durkan, Conor and Bekasov, Artur and Murray, Iain and Papamakarios, George},
	Booktitle = {Advances in Neural Information Processing Systems},
	Pages = {7509--7520},
	Title = {Neural spline flows},
	Year = {2019}}

@inproceedings{chen2018neural,
	Author = {Chen, Tian Qi and Rubanova, Yulia and Bettencourt, Jesse and Duvenaud, David K},
	Booktitle = {Advances in neural information processing systems},
	Pages = {6571--6583},
	Title = {Neural ordinary differential equations},
	Year = {2018}}

@article{LiEtAl_ScalableGradientSDE,
	Author = {Xuechen Li and Ting-Kam Leonard Wong and Ricky T. Q. Chen and David Duvenaud},
	Journal = {arXiv preprint arXiv:2001.01328},
	Title = {Scalable Gradients for Stochastic Differential Equations},
	Year = {2020}}

@article{noe2019boltzmann,
	Author = {No{\'e}, Frank and Olsson, Simon and K{\"o}hler, Jonas and Wu, Hao},
	Journal = {Science},
	Number = {6457},
	Pages = {eaaw1147},
	Publisher = {American Association for the Advancement of Science},
	Title = {Boltzmann generators: Sampling equilibrium states of many-body systems with deep learning},
	Volume = {365},
	Year = {2019}}

@article{kohler2019equivariant,
	Author = {K{\"o}hler, Jonas and Klein, Leon and No{\'e}, Frank},
	Journal = {arXiv preprint arXiv:1910.00753},
	Title = {Equivariant Flows: sampling configurations for multi-body systems with symmetric energies},
	Year = {2019}}

@article{rezende2015variational,
	Author = {Rezende, Danilo Jimenez and Mohamed, Shakir},
	Journal = {arXiv preprint arXiv:1505.05770},
	Title = {Variational inference with normalizing flows},
	Year = {2015}}

\clearpage

\clearpage
\setcounter{page}{1}
\setcounter{figure}{0} \renewcommand{\thefigure}{S\arabic{figure}} 
\setcounter{table}{0} \renewcommand{\thetable}{S\arabic{table}} 

\section*{Supplementary Material}

\subsection*{1. Training normalizing flows}

\paragraph{Energy-based training and forward weight maximization.}

If the target density $\mu_{X}$ is known up to a constant $Z_{X}$,
we minimize the forward KL divergence between the generated and the
target distribution.
\begin{align}
 & \mathrm{KL}(p_{X}\parallel\mu_{X})\label{eq:KL}\\
 & =\mathbb{E}_{\mathbf{x}\sim p_{X}(\mathbf{x})}\left[\log p_{X}(\mathbf{x})-\log\mu_{X}(\mathbf{x})\right]\nonumber \\
 & =\mathbb{E}_{\mathbf{z}\sim\mu_{Z}(\mathbf{z})}\left[u_{X}(F_{ZX}(\mathbf{z}))-\Delta S_{ZX}(\mathbf{z})\right]+\mathrm{const}.\nonumber 
\end{align}
The importance weights wrt the target distribution can be computed
as:
\begin{equation}
w_{X}(\mathbf{x})= \exp\left(-u_{X}\left(F_{ZX}(\mathbf{z})\right)+u_{Z}(\mathbf{z})+\Delta S_{ZX}(\mathbf{z})\right) \propto \frac{\mu_{X}(\mathbf{x})}{p_{X}(\mathbf{x})}.\label{eq:importance_weights_NF}
\end{equation}
As $\mathbb{E}_{\mathbf{z}\sim p_{Z}(\mathbf{z})}\left[u_{Z}(\mathbf{z})\right]$
is a constant, we can equivalently minimize KL or maximize log weights:
\begin{equation}
\max\mathbb{E}_{\mathbf{z}\sim p_{Z}(\mathbf{z})}\left[\log w_{X}(\mathbf{x})\right]=\min\mathrm{KL}(p_{X}\parallel\mu_{X}),\label{eq:KL_weight_max}
\end{equation}

\paragraph{Maximum likelihood and backward weight maximization.}

The backward KL divergence $\mathrm{KL}(\mu_{X}\parallel p_{X})$
is not always tractable as $\mu_{X}(\mathbf{x})$ can be difficult
to sample from. Replacing $\mu_{X}(\mathbf{x})$ by the empirical
data distribution $\rho_{X}(\mathbf{x})$, the KL becomes a negative
log-likelihood:
\begin{align}
 & \mathrm{NLL}(\rho_{X}\parallel p_{X})\label{eq:NLL}\\
 & =\mathbb{E}_{\mathbf{x}\sim\rho_{X}(\mathbf{x})}\left[u_{Z}(F_{XZ}(\mathbf{x}))-\Delta S_{XZ}(\mathbf{x})\right]+\mathrm{const}.\nonumber 
\end{align}
Using $\mathbb{E}_{\mathbf{x}\sim\rho_{X}(\mathbf{x})}\left[-\log\rho_{X}(\mathbf{x})\right]=\mathrm{const}$
and the weights:
\[
w_{Z}(\mathbf{z})=
\exp\left(-u_{Z}\left(F_{XZ}(\mathbf{x})\right)-\log\rho_{X}(\mathbf{x})+\Delta S_{XZ}(\mathbf{x})\right)\propto
\frac{\mu_{Z}(\mathbf{z})}{p_{Z}(\mathbf{z})},
\]
maximum likelihood equals log weight maximization:
\begin{equation}
\max\mathbb{E}_{\mathbf{x}\sim\rho_{X}(\mathbf{x})}\left[\log w_{Z}(\mathbf{z})\right]=\min\mathrm{NLL}(\rho_{X}\parallel p_{X}).\label{eq:NLL_weight_max}
\end{equation}

\subsection*{2. Proof of theorem \ref{thm:unbiased_sampling} (unbiased sampling with SNF importance weights)}

Considering
\begin{eqnarray*}
\mathbb{E}_{\mu_{X}}[O] & = & \int\mu_{X}(\mathbf{x})O(\mathbf{x})\mathrm{d}\mathbf{x}\\
 & = & \iint\mu_{X}(\mathbf{x})\mathbb{P}_{b}(\mathbf{x}\to\mathbf{z})O(\mathbf{x})\mathrm{d}\mathbf{z}\mathrm{d}\mathbf{x}\\
 & = & \iint\mu_{Z}(\mathbf{z})\mathbb{P}_{f}(\mathbf{z}\to\mathbf{x})\left[\frac{\mu_{X}(\mathbf{x})\mathbb{P}_{b}(\mathbf{x}\to\mathbf{z})}{\mu_{Z}(\mathbf{z})\mathbb{P}_{f}(\mathbf{z}\to\mathbf{x})}O(\mathbf{x})\right]\mathrm{d}\mathbf{z}\mathrm{d}\mathbf{x}\\
 & = & \mathbb{E}_{f}\left[\frac{\mu_{X}(\mathbf{x})\mathbb{P}_{b}(\mathbf{x}\to\mathbf{z})}{\mu_{Z}(\mathbf{z})\mathbb{P}_{f}(\mathbf{z}\to\mathbf{x})}O(\mathbf{x})\right],
\end{eqnarray*}
where $\mathbb{E}_{f}$ denotes the expectation over forward path realizations.
In practice, we do not know the normalization constant of $\mu_{X}$
and we therefore replace $\frac{\mu_{X}(\mathbf{x})\mathbb{P}_{b}(\mathbf{x}\to\mathbf{z})}{\mu_{Z}(\mathbf{z})\mathbb{P}_{f}(\mathbf{z}\to\mathbf{x})}$
by the unnormalized path weights in Eq. (\ref{eq:acceptance_ratio}).
Then we must normalize the estimator for expectation values,
obtaining:
\[
\frac{\sum_{k=1}^{N}\mathbf{w}(\mathbf{z}_{k}\to\mathbf{x}_{k})O(\mathbf{x}_{k})}{\sum_{k=1}^{N}\mathbf{w}(\mathbf{z}_{k}\to\mathbf{x}_{k})}\stackrel{p}{\to}\mathbb{E}_{\mu}[O]
\]
which converges towards $\mathbb{E}_{\mu}[O]$ with $N\rightarrow\infty$
according to the law of large numbers.

\subsection*{3. Derivation of the deterministic layer probability ratio}

In order to work with delta distributions, we define $\delta^{\sigma}(\mathbf{x})=\mathcal{N}(\mathbf{x};\boldsymbol{0},\sigma\mathbf{I})$,
i.e. a Gaussian normal distribution with mean $\mathbf{0}$ and variance
$\sigma$ and then consider the limit $\sigma\rightarrow0^{+}$. In
the case where $\sigma>0$, by defining
\[
q_{t}^{\sigma}(\mathbf{y}_{t}\to\mathbf{y}_{t+1})=\delta^{\sigma}(\mathbf{y}_{t+1}-F_{t}(\mathbf{y}_{t})),
\]
and
\begin{eqnarray*}
\tilde{q}_{t}^{\sigma}(\mathbf{y}_{t+1}\to\mathbf{y}_{t}) & = & \frac{p_{t}(\mathbf{y}_{t})q_{t}^{\sigma}(\mathbf{y}_{t}\to\mathbf{y}_{t+1})}{\int p_{t}(\mathbf{y})q_{t}^{\sigma}(\mathbf{y}\to\mathbf{y}_{t+1})\mathrm{d}\mathbf{y}}\\
 & = & \frac{p_{t}(\mathbf{y}_{t})\delta^{\sigma}(\mathbf{y}_{t+1}-F_{t}(\mathbf{y}_{t}))}{\int p_{t}(\mathbf{y})\delta^{\sigma}(\mathbf{y}_{t+1}-F_{t}(\mathbf{y}))\mathrm{d}\mathbf{y}},
\end{eqnarray*}
we have
\[
\frac{\tilde{q}_{t}^{\sigma}(\mathbf{y}_{t+1}\to\mathbf{y}_{t})}{q_{t}^{\sigma}(\mathbf{y}_{t}\to\mathbf{y}_{t+1})}=\frac{p_{t}(\mathbf{y}_{t})}{\int p_{t}(\mathbf{y})\delta^{\sigma}(\mathbf{y}_{t+1}-F(\mathbf{y}))\mathrm{d}\mathbf{y}},
\]
where $p_{t}(y_{t})$ denotes the marginal distribution of $\mathbf{y}_{t}$.
By considering
\begin{align*}
\lim_{\sigma\to0^{+}}\int p_{t}(\mathbf{y})\delta^{\sigma}(\mathbf{y}_{t+1}-F_{t}(\mathbf{y}))\mathrm{d}\mathbf{y}
 & =\lim_{\sigma\to0^{+}}\int p_{t}(F_{t}^{-1}(\mathbf{y}'))\delta^{\sigma}(\mathbf{y}_{t+1}-\mathbf{y}')\left|\mathrm{det}\left(\frac{\partial F_{t}^{-1}(\mathbf{y}')}{\partial\mathbf{y}'}\right)\right|\mathrm{d}\mathbf{y}'\\
 & =p_{t}(F_{t}^{-1}(\mathbf{y}_{t+1}))\left|\mathrm{det}\left(\frac{\partial F_{t}^{-1}(\mathbf{y}_{t+1})}{\partial\mathbf{y}_{t+1}}\right)\right|\\
 & =p_{t}(\mathbf{y}_{t})\left|\det\mathbf{J}_{t}(\mathbf{y}_{t})\right|^{-1}
\end{align*}
and using the definition of $\Delta S_{t}$ in terms of path probability
rations, we obtain:
\begin{align*}
 \exp\left(\Delta S_{t}\right)
 & =\frac{\tilde{q}_{t}(\mathbf{y}_{t+1}\to\mathbf{y}_{t})}{q_{t}(\mathbf{y}_{t}\to\mathbf{y}_{t+1})}=\lim_{\sigma\rightarrow0^{+}}\frac{\tilde{q}_{t}^{\sigma}(\mathbf{y}_{t+1}\to\mathbf{y}_{t})}{q_{t}^{\sigma}(\mathbf{y}_{t}\to\mathbf{y}_{t+1})}\\
 & =\lim_{\sigma\rightarrow0^{+}}\frac{p_{t}(\mathbf{y}_{t})}{\int p_{t}(\mathbf{y})\delta^{\sigma}(\mathbf{y}_{t+1}-F(\mathbf{y}))\mathrm{d}\mathbf{y}}\\
 & =\left|\det\mathbf{J}_{t}(\mathbf{y}_{t})\right|
\end{align*}
and thus
\[
\Delta S_{t}=\log\left|\det\mathbf{J}_{t}(\mathbf{y}_{t})\right|.
\]

\subsection*{4. Derivation of the overdamped Langevin path probability ratio}

These results follow \cite{NilmeyerEtAl_PNA11_NCMC}. The backward
step is realized by 
\begin{equation}
\mathbf{y}_{t}=\mathbf{y}_{t+1}-\epsilon_{t}\nabla u_{\lambda}(\mathbf{y}_{t+1})+\sqrt{\frac{2\epsilon}{\beta}}\tilde{\boldsymbol{\eta}}_{t}.\label{eq:BD_dynamics_rev}
\end{equation}
Combining Equations (\ref{eq:BD_dynamics1}) and (\ref{eq:BD_dynamics_rev}):
\begin{align*}
-\epsilon_{t}\nabla u_{\lambda}(\mathbf{y}_{t})+\sqrt{\frac{2\epsilon_{t}}{\beta}}\boldsymbol{\eta}_{t} & =\epsilon_{t}\nabla u_{\lambda}(\mathbf{y}_{t+1})-\sqrt{\frac{2\epsilon_{t}}{\beta}}\tilde{\boldsymbol{\eta}}_{t}.
\end{align*}
and thus
\[
\tilde{\boldsymbol{\eta}}_{t}=\sqrt{\frac{\epsilon_{t}\beta}{2}}\left[\nabla u_{\lambda}(\mathbf{y}_{t})+\nabla u_{\lambda}(\mathbf{y}_{t+1})\right]-\boldsymbol{\eta}_{t}.
\]
Resulting in the path probability ratio:
\begin{align*}
\exp\left(\Delta S_{t}\right) & =\frac{q_{t}(\mathbf{y}_{t+1}\to\mathbf{y}_{t})}{q_{t}(\mathbf{y}_{t}\to\mathbf{y}_{t+1})}=\frac{p(\tilde{\boldsymbol{\eta}}_{t})\left|\frac{\partial\mathbf{y}_{t}}{\partial\tilde{\boldsymbol{\eta}}_{t}}\right|}{p(\boldsymbol{\eta}_{t})\left|\mathrm{det}\left(\frac{\partial\mathbf{y}_{t+1}}{\partial\boldsymbol{\eta}_{t}}\right)\right|}\\
 & =\frac{p(\tilde{\boldsymbol{\eta}}_{t})}{p(\boldsymbol{\eta}_{t})}=\mathrm{e}^{-\frac{1}{2}\left(\left\Vert \tilde{\boldsymbol{\eta}}_{t}\right\Vert ^{2}-\left\Vert \boldsymbol{\eta}_{t}\right\Vert ^{2}\right)}.
\end{align*}
and thus
\[
-\Delta S_{t}=\frac{1}{2}\left(\left\Vert \tilde{\boldsymbol{\eta}}_{t}\right\Vert ^{2}-\left\Vert \boldsymbol{\eta}_{t}\right\Vert ^{2}\right)
\]

\subsection*{5. Derivation of the Langevin probability ratio}

These results follow \cite{NilmeyerEtAl_PNA11_NCMC}. We define constants:
\begin{align*}
c_{1} & =\frac{\Delta t}{2m}\\
c_{2} & =\sqrt{\frac{4\gamma m}{\Delta t\beta}}\\
c_{3} & =1+\frac{\gamma\Delta t}{2}
\end{align*}
Then, the forward step of Brooks-Brünger-Karplus (BBK, leap-frog)
Langevin dynamics are defined as:
\begin{align}
\mathbf{v}' & =\mathbf{v}_{t}+c_{1}\left[-\nabla u_{\lambda}(\mathbf{x}_{t})-\gamma m\mathbf{v}_{t}+c_{2}\boldsymbol{\eta}_{t}\right]\label{eq:BBK_Apx1}\\
\mathbf{x}_{t+1} & =\mathbf{x}_{t}+\Delta t\mathbf{v}'\label{eq:BBK_Apx2}\\
\mathbf{v}_{t+1} & =\frac{1}{c_{3}}\left[\mathbf{v}'+c_{1}\left(-\nabla u_{\lambda}(\mathbf{x}_{t+1})+c_{2}\boldsymbol{\eta}_{t}'\right)\right]\label{eq:BBK_Apx3}
\end{align}
Note that the factor $4$ in sqrt is different from \cite{NilmeyerEtAl_PNA11_NCMC}
-- this factor is needed as we employ $\Delta t/2$ in both half-steps.
The backward step with reversed momenta, $(\mathbf{x}_{t+1},-\mathbf{v}_{t+1})\rightarrow(\mathbf{x}_{t},-\mathbf{v}_{t})$
is then defined by:
\begin{align}
\mathbf{v}'' & =-\mathbf{v}_{t+1}+c_{1}\left[-\nabla u_{\lambda}(\mathbf{x}_{t+1})+\gamma m\mathbf{v}_{t+1}+c_{2}\tilde{\boldsymbol{\eta}}_{t}\right]\label{eq:BBK_Apx4}\\
\mathbf{x}_{t} & =\mathbf{x}_{t+1}+\Delta t\mathbf{v}''\label{eq:BBK_Apx5}\\
-\mathbf{v}_{t} & =\frac{1}{c_{3}}\left[\mathbf{v}''+c_{1}\left(-\nabla u_{\lambda}(\mathbf{x}_{t})+c_{2}\tilde{\boldsymbol{\eta}}_{t}'\right)\right]\label{eq:BBK_Apx6}
\end{align}
To compute the momenta $\tilde{\boldsymbol{\eta}}_{t},\tilde{\boldsymbol{\eta}}_{t}'$
that realize the reverse step, we first combine Eqs. (\ref{eq:BBK_Apx2}-\ref{eq:BBK_Apx5})
to obtain:
\begin{equation}
\mathbf{v}'=-\mathbf{v}''\label{eq:BBK_Apx7}
\end{equation}
Combining Eqs. (\ref{eq:BBK_Apx3}), (\ref{eq:BBK_Apx4}) and (\ref{eq:BBK_Apx7}),
we obtain:
\begin{align*}
\left(1+\frac{\gamma\Delta t}{2}\right)\mathbf{v}_{t+1} & =\mathbf{v}'+c_{1}\left(-\nabla u_{\lambda}(\mathbf{x}_{t+1})+c_{2}\boldsymbol{\eta}_{t}'\right)\\
\left(1-\frac{\gamma\Delta t}{2}\right)\mathbf{v}_{t+1} & =\mathbf{v}'+c_{1}\left(-\nabla u_{\lambda}(\mathbf{x}_{t+1})+c_{2}\tilde{\boldsymbol{\eta}}_{t}\right),
\end{align*}
and:
\begin{align*}
\tilde{\boldsymbol{\eta}}_{t} & =\boldsymbol{\eta}_{t}'-\sqrt{\gamma\Delta tm\beta}\mathbf{v}_{t+1}
\end{align*}
Combining Eqs. (\ref{eq:BBK_Apx1}), (\ref{eq:BBK_Apx6}) and (\ref{eq:BBK_Apx7}),
we obtain:
\begin{align*}
-\mathbf{v}_{t}\left(1-\frac{\gamma\Delta t}{2}\right) & =\mathbf{v}''+c_{1}\left(-\nabla u_{\lambda}(\mathbf{x}_{t})+c_{2}\boldsymbol{\eta}_{t}\right)\\
-\mathbf{v}_{t}\left(1+\frac{\gamma\Delta t}{2}\right) & =\mathbf{v}''+c_{1}\left(-\nabla u_{\lambda}(\mathbf{x}_{t})+c_{2}\tilde{\boldsymbol{\eta}}_{t}'\right),
\end{align*}
and:
\begin{align*}
-\mathbf{v}_{t}\left(1-\frac{\gamma\Delta t}{2}\right)-c_{2}\boldsymbol{\eta}_{t} & =-\mathbf{v}_{t}\left(1+\frac{\gamma\Delta t}{2}\right)-c_{2}\tilde{\boldsymbol{\eta}}_{t}'\\
\tilde{\boldsymbol{\eta}}_{t}' & =\boldsymbol{\eta}_{t}-\sqrt{\gamma\Delta tm\beta}\mathbf{v}_{t}
\end{align*}
To compute the path probability ratio we introduce the Jacobian
\[
J(\boldsymbol{\eta}_{t},\boldsymbol{\eta}_{t}')=\det\left[\begin{array}{cc}
\frac{\partial\mathbf{x}_{t+1}}{\partial\boldsymbol{\eta}_{t}} & \frac{\partial\mathbf{v}_{t+1}}{\partial\boldsymbol{\eta}_{t}}\\
\frac{\partial\mathbf{x}_{t+1}}{\partial\boldsymbol{\eta}_{t}'} & \frac{\partial\mathbf{v}_{t+1}}{\partial\boldsymbol{\eta}_{t}'}
\end{array}\right]
\]
and find:
\begin{align*}
\exp\left(\Delta S_{t}\right) & =\frac{\tilde{q}_{t}\left((\mathbf{x}_{t+1},-\mathbf{v}_{t+1})\rightarrow(\mathbf{x}_{t},\mathbf{v}_{t})\right)}{q_{t}\left((\mathbf{x}_{t},\mathbf{v}_{t})\rightarrow(\mathbf{x}_{t+1},-\mathbf{v}_{t+1})\right)}\\
 & =\frac{p(\tilde{\boldsymbol{\eta}}_{t})p(\tilde{\boldsymbol{\eta}}_{t}')J(\tilde{\boldsymbol{\eta}}_{t},\tilde{\boldsymbol{\eta}}_{t}')}{p(\boldsymbol{\eta}_{t})p(\boldsymbol{\eta}_{t}')J(\boldsymbol{\eta}_{t},\boldsymbol{\eta}_{t}')}\\
-\Delta S_{t} & =\frac{1}{2}\left(\left(\left\Vert \tilde{\boldsymbol{\eta}}_{t}\right\Vert ^{2}+\left\Vert \tilde{\boldsymbol{\eta}}_{t}'\right\Vert ^{2}\right)-\left(\left\Vert \boldsymbol{\eta}_{t}\right\Vert ^{2}+\left\Vert \boldsymbol{\eta}_{t}'\right\Vert ^{2}\right)\right)
\end{align*}
where the Jacobian ratio cancels as the Jacobians are independent
of the noise variables.

\subsection*{6. Derivation of the probability ratio for Markov Chain Monte Carlo}

For MCMC, $q_{t}$ satisfies the detailed balance condition
\[
\exp(-u_{\lambda}(\mathbf{y}_{t}))\cdot q_{t}(\mathbf{y}_{t}\to\mathbf{y}_{t+1})=\exp(-u_{\lambda}(\mathbf{y}_{t+1}))\cdot\tilde{q}_{t}(\mathbf{y}_{t+1}\to\mathbf{y}_{t})
\]
with respect to the potential function $u_{\lambda}$. We have
\begin{eqnarray*}
\Delta S_{t} & = & \log\frac{\tilde{q}_{t}(\mathbf{y}_{t+1}\to\mathbf{y}_{t})}{q_{t}(\mathbf{y}_{t}\to\mathbf{y}_{t+1})}\\
 & = & \log\frac{\exp(-u_{\lambda}(\mathbf{y}_{t}))}{\exp(-u_{\lambda}(\mathbf{y}_{t+1}))}\\
 & = & u_{\lambda}(\mathbf{y}_{t+1})-u_{\lambda}(\mathbf{y}_{t})
\end{eqnarray*}

\subsection*{7. Derivation of the probability ratio for Hamiltonian MC with Metropolis
acceptance}

Hamiltonian MC with Metropolis acceptance defines a forward path density
\[
q_{t}\left((\mathbf{y}_{t},\mathbf{v})\to(\mathbf{y}_{t+1},\mathbf{v}^{K})\right)
\]
which satisfies the joint detailed balance condition
\begin{align}
 & \exp(-u_{\lambda}(\mathbf{y}_{t}))\mathcal{N}(\mathbf{v}|\mathbf{0},\mathbf{I})\cdot q_{t}\left((\mathbf{y}_{t},\mathbf{v})\to(\mathbf{y}_{t+1},\mathbf{v}^{K})\right)\nonumber \\
 & \quad=\exp(-u_{\lambda}(\mathbf{y}_{t+1}))\mathcal{N}(\mathbf{v}^{K}|\mathbf{0},\mathbf{I})\cdot\tilde{q}_{t}\left((\mathbf{y}_{t+1},\mathbf{v}^{K})\to(\mathbf{y}_{t},\mathbf{v})\right).\label{eq:hmc-db}
\end{align}
Considering the velocity $\mathbf{v}$ is independently drawn from
$\mathcal{N}(\mathbf{v}|\mathbf{0},\mathbf{I})$, the ``marginal''
forward path density of $\mathbf{y}_{t}\to\mathbf{y}_{t+1}$ is
\[
q_{t}\left(\mathbf{y}_{t}\to\mathbf{y}_{t+1}\right)=\iint\mathcal{N}(\mathbf{v}|\mathbf{0},\mathbf{I})\cdot q_{t}\left((\mathbf{y}_{t},\mathbf{v})\to(\mathbf{y}_{t+1},\mathbf{v}^{K})\right)\mathrm{d}\mathbf{v}\mathrm{d}\mathbf{v}^{K}.
\]
Then, it can be obtained from \eqref{eq:hmc-db} that
\begin{align*}
\exp(-u_{\lambda}(\mathbf{y}_{t}))q_{t}\left(\mathbf{y}_{t}\to\mathbf{y}_{t+1}\right)
= & \iint\exp(-u_{\lambda}(\mathbf{y}_{t}))\mathcal{N}(\mathbf{v}|\mathbf{0},\mathbf{I})\left((\mathbf{y}_{t},\mathbf{v})\to(\mathbf{y}_{t+1},\mathbf{v}^{K})\right)\mathrm{d}\mathbf{v}\mathrm{d}\mathbf{v}^{K}\\
= & \iint\exp(-u_{\lambda}(\mathbf{y}_{t+1}))\mathcal{N}(\mathbf{v}^{K}|\mathbf{0},\mathbf{I})\tilde{q}_{t}\left((\mathbf{y}_{t+1},\mathbf{v}^{K})\to(\mathbf{y}_{t},\mathbf{v})\right)\mathrm{d}\mathbf{v}\mathrm{d}\mathbf{v}^{K}\\
= & \exp(-u_{\lambda}(\mathbf{y}_{t+1}))\tilde{q}_{t}\left(\mathbf{y}_{t+1}\to\mathbf{y}_{t}\right),
\end{align*}
and
\begin{align*}
\Delta S_{t} & =\log\frac{\tilde{q}_{t}\left(\mathbf{y}_{t+1}\to\mathbf{y}_{t}\right)}{q_{t}\left(\mathbf{y}_{t}\to\mathbf{y}_{t+1}\right)}\\
 & =u_{\lambda}(\mathbf{y}_{t+1})-u_{\lambda}(\mathbf{y}_{t})
\end{align*}

\subsection*{8. Details on using SNFs for variational inference}

Here we elaborate on the details of using SNFs as a variational approximation of the posterior distribution of a \textit{variational autoencoder} (VAE) \cite{KingmaWelling_ICLR14_VAE} as presented in our last results section. In contrast to the usual notation used in common VAE literature, we choose $\mathbf{x}$ to indicate the \textit{latent} variable, while we call the \textit{observed} variable $\mathbf{s}$. This is due to being consistent with the use of $\mathbf{x}$ as the sampled variable of interest throughout our former discussions.

For a given data set $ \{\mathbf{s}_{1},\ldots,\mathbf{s}_{N}\}$,
the decoder $D$ of a VAE characterizes each $\mathbf{s}$ as a random
variable with a tractable distribution $\mathbb{P}_{D}(\mathbf{s}|\mathbf{x})$
depending on a unknown latent variable $\mathbf{x}$. Furthermore, the prior
distribution is assumed to be tractable as well (e.g. an isotropic normal
distribution). Here we take the prior
\begin{align*}
    \mathbb{P}(\mathbf{x}) = \mathcal{N}(\mathbf{x}\mid\mathbf{0},\mathbf{I}).
\end{align*}
Together, this defines the joint distribution
\[
\mathbb{P}_{D}(\mathbf{x},\mathbf{s})= \mathbb{P}(\mathbf{x}) \cdot\mathbb{P}_{D}(\mathbf{s}|\mathbf{x}).
\]
Conditioned on a given $\mathbf{s}$, we can utilize a SNF to approximate the
posterior distribution
\[
\mathbb{P}_{D}(\mathbf{x}|\mathbf{s})=\frac{\mathbb{P}_{D}(\mathbf{x},\mathbf{s})}{\mathbb{P}_{D}(\mathbf{s})}.
\]
For convenience and consistency with the former discussion, we define $\mu_{X}(\mathbf{x})=\mathbb{P}_{D}(\mathbf{x}|\mathbf{s})$
and $u_{X}(\mathbf{x})=-\log\mathbb{P}_{D}(\mathbf{x},\mathbf{s})$.
Thus, the parameters of the SNF and the decoder $D$ can be trained by minimizing $J_{KL}$  which provides an
upper bound of the negative log-likelihood of $\mathbf{s}$ as follows:
\begin{eqnarray*}
J_{KL}(\mathbf{s}) & = & \mathbb{E}_{\mathbf{z}\sim\mu_{Z},\mathbf{y}_{1},\ldots,\mathbf{y}_{T}}[u_{X}(\mathbf{y}_{T})-\sum_{t=0}^{T-1}\Delta S_{t}]\\
 & = & \mathbb{E}_{\mathbf{z}\sim\mu_{Z},\mathbf{y}_{1},\ldots,\mathbf{y}_{T}}[-\log\mu_{X}(\mathbf{y}_{T})-\sum_{t=0}^{T-1}\Delta S_{t}] -\log\mathbb{P}_{D}(\mathbf{s})\\
 & = & \mathrm{KL}(\mu_{Z}(\mathbf{z})\mathbb{P}_{f}(\mathbf{z}\to\mathbf{x})\parallel\mu_{X}(\mathbf{x})\mathbb{P}_{b}(\mathbf{x}\to\mathbf{z})) -\log\mathbb{P}_{D}(\mathbf{s})\\
 & \ge & -\log\mathbb{P}_{D}(\mathbf{s})
\end{eqnarray*}
If the SNF consists of only deterministic transformations, $J_{KL}$
is equivalent to $\mathcal{F}$ in \cite{rezende2015variational}.

We estimate $J_{KL}(\mathbf{s})$ on samples as
\begin{equation}
\hat{J}_{KL}(\mathbf{s})=\frac{1}{M}\sum_{i=1}^{M}u_{X}(\mathbf{y}_{T}^{(i)})-\sum_{t=0}^{T-1}\Delta S_{t}^{(i)},\label{eq:JKL-vi}
\end{equation}
by sampling $M$ paths $\{(\mathbf{y}_{0}^{(i)},\ldots,\mathbf{y}_{T}^{(i)})\}_{i=1}^{M}$
for each $\mathbf{s}$ and setting $M=5$. 

\paragraph{Estimating the evidence.}
After training we approximate
$-\log\mathbb{P}_{D}(\mathbf{s})$ by marginalizing out the latent variable $\mathbf{x}$ via Monte Carlo sampling. 
In order to improve sampling efficiency and have a fair comparison among the three different SNF instantiations, we approximate the posterior distribution $\mathbb{P}_{D}(\mathbf{x} | \mathbf{s})$ of the trained model using the same  variational approximation:

\begin{enumerate}
\item We define a simple base distribution $q(\mathbf{z}) = \mathcal{N}(\mathbf{z}|\mathbf{0}, \mathbf{I})$, together with a conditional diffeomorphism $F_{LL}(\mathbf{x}|\mathbf{s})$ transforming $\mathbf{z}$ to $\mathbf{x}$ and vice-versa conditioned on $\mathbf{s}$:
\[
\begin{array}{lcccr}
   & & F^{-1}_{LL}(\cdot|\mathbf{s}) &  &\\
   & \mathbf{x} & \rightleftarrows & \mathbf{z}. \\
   &  &F_{LL}(\cdot|\mathbf{s}) & & \\
\end{array}
\]
We realize such a conditional flow via RealNVP transformations, where coupling layers are additionally conditioned on $\mathbf{s}$ and only $\mathbf{x} / \mathbf{z}$ is transformed during the flow. Together with $q(\mathbf{z})$ this defines the conditional distribution 
\begin{align*}
    q_{LL}(\mathbf{x} | \mathbf{s}) = q(F_{LL}^{-1}(\mathbf{x}|\mathbf{s})) \left|\mathrm{det}\left( \frac{\partial F_{LL}^{-1}(\mathbf{x}|\mathbf{s})}{\partial \mathbf{x}} \right)\right|
\end{align*}
which we use as variational approximation to the true posterior. We then train $q_{LL}$ by minimizing the KL divergence
\begin{align*}
    \mathbb{E}_{\mathbf{s}, \mathbf{z} \sim q(\mathbf{z})}\left[
         \log q_{LL}(F_{LL}(\mathbf{z}|\mathbf{s})|\mathbf{s})
 -\log\mathbb{P}(F_{LL}(\mathbf{z}|\mathbf{s}))-\log\mathbb{P}_{D}(\mathbf{s}|F_{LL}(\mathbf{z}|\mathbf{s})) \right] + const.
\end{align*}
until convergence. This loss is minimized iff $q_{LL}(\mathbf{x} | \mathbf{s}) = \mathbb{P}_{D}(\mathbf{x} | \mathbf{s})$.
    
\item Now considering
\begin{align*}
\mathbb{P}_{D}(\mathbf{s})
= & \int \mathbb{P}(\mathbf{x}) \mathbb{P}_{D}(\mathbf{s}|\mathbf{x})\mathrm{d}\mathbf{x}\\
= & \int q_{LL}(\mathbf{x}|\mathbf{s})\frac{\mathbb{P}(\mathbf{x})\mathbb{P}_{D}(\mathbf{s}|\mathbf{x})}{q_{LL}(\mathbf{x}|\mathbf{s})}\mathrm{d}\mathbf{x}\\
= & \mathbb{E}_{\mathbf{x}\sim q_{LL}(\mathbf{x}|\mathbf{s})}\left[\frac{\mathbb{P}(\mathbf{x})\mathbb{P}_{D}(\mathbf{s}|\mathbf{x})}{q_{LL}(\mathbf{x}|\mathbf{s})}\right]\\
= & \mathbb{E}_{\mathbf{z}\sim\mathcal{N}(\mathbf{z}|\mathbf{0},\mathbf{I})}\left[\frac{\mathbb{P}(F_{LL}(\mathbf{z}|\mathbf{s}))\mathbb{P}_{D}(\mathbf{s}|F_{LL}(\mathbf{z}|\mathbf{s}))}{q(\mathbf{z})\left|\mathrm{det}\left( \frac{\partial F_{LL}^{-1}(\mathbf{x}|\mathbf{s})}{\partial \mathbf{x}} \right)\right|^{-1}}\right],
\end{align*}
we can draw $N$ samples $\mathbf{z}^{(1)},\ldots,\mathbf{z}^{(N)}$
and approximate $\mathbb{P}_{D}(\mathbf{s})$ by
\[
\hat{\mathbb{P}}_{D}(\mathbf{s})=\frac{1}{N}\sum_{i=1}^{N}\frac{\mathbb{P}(F_{LL}(\mathbf{z^{(i)}}|\mathbf{s}))\mathbb{P}_{D}(\mathbf{s}|F_{LL}(\mathbf{z}^{(i)}|\mathbf{s}))}{q(\mathbf{z}^{(i)})\left|\mathrm{det}\left( \frac{\partial F_{LL}^{-1}(\mathbf{x}|\mathbf{s})}{\partial \mathbf{x}} \right)\right|^{-1}}.
\]
In experiments, $N$ is set to be $2000$.
\end{enumerate}
In table \ref{tab:VAE}, the first column is the mean value of $\hat{J}_{KL}(\mathbf{s})$
on the test data set as a variational bound of the mean of $-\log p(\mathbf{s})$
(related to Fig. 4a in \cite{rezende2015variational}). The second
column is the mean value of $-\log\hat{\mathbb{P}}_{D}(\mathbf{s})$
on the test data set (related to Fig. 4c in \cite{rezende2015variational}).

\subsection*{9. Hyper-parameters and other benchmark details}

All experiments were run using PyTorch 1.2 and on GTX1080Ti cards.
Optimization uses Adam \cite{KingmaBa_ADAM} with step-size $0.001$
and otherwise default parameters. All deterministic flow transformations
use RealNVP \cite{DinhBengio_RealNVP}. A RealNVP block is defined
by two subsequent RealNVP layers that are swapped such that each channel
gets transformed once as a function of the other channel. The affine transformation of each RealNVP layer is given by a fully connected ReLU
network.
For the NSF layers we substitute the simple affine transformations used in RealNVP by the rational-quadratic (RQ) spline transformation implemented in \texttt{https://github.com/bayesiains/nflows}. As before the width, height and slope of the RQ transformations are given by fully connected ReLU networks.
Again a NSF block consists of two subsequent NSF layers with intermediate swap layers.

\paragraph{Double well examples in Figures \ref{fig:double_well_illustration}
and \ref{fig:reweighting_2well}}
\begin{itemize}
\item Both normalizing flow and SNF networks use 3 RealNVP blocks with three hidden layers of dimension 64. The SNF
additionally uses 20 Metropolis MC steps per block using a Gaussian
proposal density with standard deviation 0.25.
\item Training is done by minimizing $J_{ML}$ for 300 iterations and $\frac{1}{2}J_{ML}+\frac{1}{2}KL$
for 300 iterations using a batch-size of 128.
\item ``Biased data'' is defined by running local Metropolis MC in each
of the two wells. These simulations do not transition to the other
well and we use 1000 data points in each well for training.
\item ``Unbiased data'' is produced by running Metropolis MC with a large
proposal step (standard deviation 1.5) to convergence and retaining
10000 data points for training.
\item In Table \ref{tab:adaptive_mc}, the sampling results of SNFs with RealNVP blocks and Metropolis MC steps. MC step sizes of the first SNF is fixed to be $0.25$ as before, and all step sizes of the second one are trainable parameters in $[0.01,0.3]$. The other settings are the same as in Table \ref{tab:reweighting_2well}.
\end{itemize}

\begin{table}[h]
\caption{Unbiased sampling for double well potential by SNFs with nontrainable/trainable MC step sizes.}
\label{tab:adaptive_mc}
    \centering
    \begin{tabular}{lcccccc}
     \toprule 
  & \multicolumn{3}{c}{not reweighted} & \multicolumn{3}{c}{reweighted}\\ 
  & bias & $\sqrt{\textrm{var}}$ & $\sqrt{\textrm{bias}^2 \hspace{-0.1cm}+\hspace{-0.1cm} \textrm{var}}$ & bias & $\sqrt{\textrm{var}}$ & $\sqrt{\textrm{bias}^2 \hspace{-0.1cm}+\hspace{-0.1cm} \textrm{var}}$\\ 
\midrule
RNVP + MC & $1.5\pm0.2$  & $0.3\pm0.1$ & $1.5\pm0.2$ & $0.2\pm0.1$ & $0.6\pm0.1$ & $0.6\pm0.1$\\
\tabincell{l}{\textbf{RNVP + MC with}\\\textbf{trainable step sizes}}  & $1.0\pm0.2$ & $0.2\pm0.1$ & $1.0\pm0.2$ & $0.1\pm0.1$ & $0.4\pm0.1$ & $0.4\pm0.1$ \\

\bottomrule
\end{tabular}
\end{table}

\textbf{Two-dimensional image densities in Figure \ref{fig:Images}}
\begin{itemize}
\item RealNVP and NSF flows both use 5 blocks. 
All involved transformation parameters (translation/scale in RealNVP layers, width/height/slope in NSF layers) use three hidden layers of dimension 64.
For the NSF layers we used 20 knot points in the RQ-spline transformation.
Training
was done by minimizing $J_{ML}$ for 2000 iterations with batch-size
250.
\item Purely stochastic flow (column 2) uses five blocks with 10 Metropolis
MC steps each using a Gaussian proposal density with standard deviation
0.1.
\item SNF (column 3/5) uses 5 blocks (RNVP/NSF block and 10 Metropolis MC steps
with same parameters as above). Training was done by minimizing $J_{ML}$
for 6000 iterations with batch-size 250.
\end{itemize}

\paragraph{Alanine dipeptide in Fig. \ref{fig:Ala2}}
\begin{itemize}
\item Normalizing flow uses 3 RealNVP blocks with 3 hidden layers and $[128,128,128]$
nodes in their transformers. Training was done by minimizing $J_{ML}$
for 1000 iterations with batch-size 256.
\item SNF uses the same architecture and training parameters, but additionally
20 Metropolis MC steps each using a Gaussian proposal density with
standard deviation 0.1.
\item As a last flow layer before $\mathbf{x}$, we used an invertible transformation
between Cartesian coordinates and internal coordinates (bond lengths,
angles, torsion angles) following the procedure described in \cite{noe2019boltzmann}.
The internal coordinates were normalized by removing the mean and
dividing by the standard deviation of their values in the training
data.
\item Training data: We set up Alanine dipeptide in vacuum using \href{https://openmmtools.readthedocs.io/en/0.18.1/}{OpenMMTools}.
Parameters are defined by the force field ff96 of the AMBER program
\cite{amber}. Simulations are run at standard OpenMMTools parameters
with no bond constraints, 1 femtosecond time-step for $10^{6}$ time-steps
(1 nanosecond) at a temperature of 1000 K in order to facilitate
rapid exploration of the $\phi/\psi$ torsion angles and a few hundred transitions between metastable states. $10^{5}$ atom positions were saved as
training data.
\end{itemize}

\paragraph{MNIST and Fashion-MNIST VAE in Table \ref{tab:VAE}}

\begin{itemize}
\item The latent space dimension was set to 50. The decoder consists of
2 fully connected hidden layers, with 1024 units and ReLU non-linearities
for each hidden layer. The activation function of the the output layer
is sigmoid function. $\mathbb{P}_{D}(\mathbf{s}|\mathbf{x})$ is defined
as
\begin{align*}
 & \log\mathbb{P}_{D}(\mathbf{s}|\mathbf{x})=\log\mathbb{P}_{D}(\mathbf{s}|D(\mathbf{x}))\\
 & =\sum_{i=1}^{784}[\mathbf{s}]_{i}\log[D(\mathbf{x})]_{i}+\left(1-[\mathbf{s}]_{i}\right)\log\left(1-[D(\mathbf{x})]_{i}\right),
\end{align*}
where $[\mathbf{s}]_{i}$, $[D(\mathbf{x})]_{i}$ denote the $i$th
pixel of $\mathbf{s}$ and the $i$th output of $D$.
\item Adam algorithm is used to train all models. Training was done by minimizing
$\hat{J}_{KL}$ (see \eqref{eq:JKL-vi}) for 40 epochs with batch-size
128 and step size $10^{-3}$ unless otherwise stated.
\item In simple VAE, the encoder $E$ consists of 2 fully connected hidden
layers, with 1024 nodes and ReLU non-linearities for each hidden layer.
The encoder has $100$ outputs, where the activation function of the
first $50$ outputs is the linear function and the activation function
of the last $50$ outputs is the absolute value function. The transformation
from $\mathbf{z}$ to $\mathbf{x}$ is given by
\begin{equation}
[\mathbf{x}]_{i}=[E(\mathbf{s})]_{i}+[\mathbf{z}]_{i}\cdot[E(\mathbf{s})]_{i+50}.\label{eq:encoder}
\end{equation}
\item MCMC uses 30 Metropolis MC steps each using a overdamped Langevin
proposal, where the interpolated potential are used. The interpolation
coefficients and the step size of the proposal are both trained as
parameters of the flow.
\item Normalizing flow uses 6 RealNVP blocks with 2 hidden layers and $[64,64]$
nodes in their transformers.
\item SNF uses three units with each unit consisting of 2 RealNVP blocks
+ 10 Metropolis MC steps, where architectures are the same as the
above. During the training procedure, we first train parameters of
the 6 RealNVP blocks without the Metropolis MC steps for 20 epochs,
and then train all parameters for another 20 epochs. The training
step size is $10^{-3}$ for the first 20 epochs and $10^{-4}$ for
the last 20 epochs.
\item For calculating the marginal likelihood $\mathbb{P}_{D}(\mathbf{s})$,
$F_{LL}$ uses 12 RealNVP blocks with 2 hidden layers and $[256,256]$
nodes in their transformers.
\end{itemize}

\subsection*{10. Comparison with related sampling methods}
A brief comparison of the proposed SNF and selected sampling methods with learnable proposals and transformations is provided in Table \ref{tab:method_comparison}.
Most previous sampling methods are developed based on the detailed balance in each step, except that HVI presented in \cite{salimans2015markov} can perform nonequilibrium sampling steps by using annealed target distributions.
Furthermore, some sampling techniques \cite{Hoffman_ICML17_LatentGaussianModels}
and \cite{hoffman2019neutra} also improve the sampling efficiency by linear or nonlinear deterministic transformation, where the transformation is performed only once.
It can be seen from the comparison that SNF provides a universal framework for sampling, where the deterministic and stochastic blocks can be flexibly designed and combined.

\begin{table}[h]
\caption{A comparison of samplers with learnable proposals/transformations.}
    \centering
    \label{tab:method_comparison}
    \begin{tabular}{ccc}
    \toprule
    {Method} & {\tabincell{c}{Containing nonequilibrium\\sampling steps}} & {\tabincell{c}{Combining with learnable\\deterministic transformation}}\\
    \midrule
    HVI \cite{salimans2015markov} & {$\checkmark$} & {$\times$}\\
    L2HMC \cite{levy2017generalizing} & {$\times$} & {$\times$}\\
    A-NICE-MC \cite{SongEtAl_NIPS17_ANiceMC} & {$\times$} & {$\times$}\\
    HMC for DLGMs \cite{Hoffman_ICML17_LatentGaussianModels} & {$\times$} & {$\checkmark$}\\
    NeuTra \cite{hoffman2019neutra} & {$\times$} & {$\checkmark$}\\
    SNF & {$\checkmark$} & {$\checkmark$}\\
    \bottomrule
    \end{tabular}
\end{table}

\clearpage

\subsection*{Supplementary Figures}

\begin{figure}[H]
\includegraphics[width=1\columnwidth]{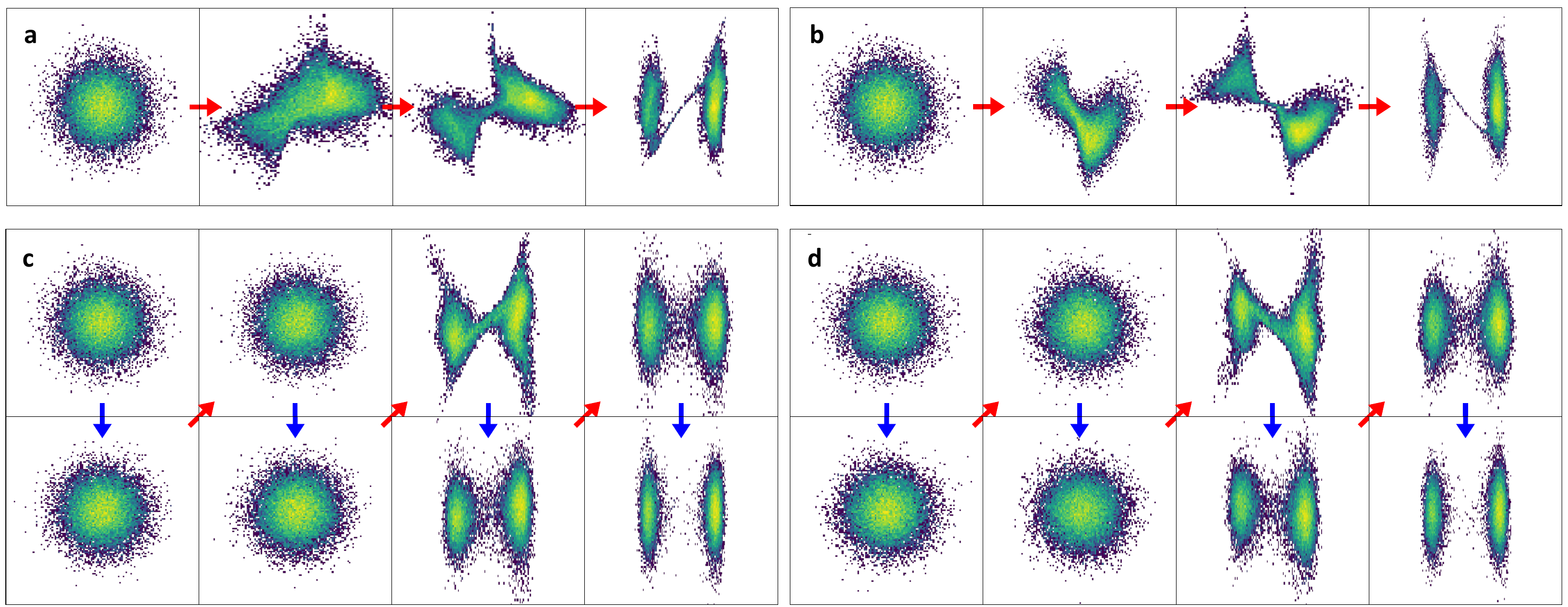}

\caption{\label{fig:SI_2well_Reproducibility}\textbf{Reproducibility of normalizing
flows for the double well}. Red arrows indicate deterministic transformations
(perturbations), blue arrows indicate stochastic dynamics (relaxations).
\textbf{a-b}) Two independent runs of 3 RealNVP blocks (6 layers).
\textbf{c-d}) Two independent runs of same architecture with 20 BD
steps before/after RealNVP blocks.}
\end{figure}

\begin{figure}[H]
\begin{centering}
\includegraphics[width=0.4\columnwidth]{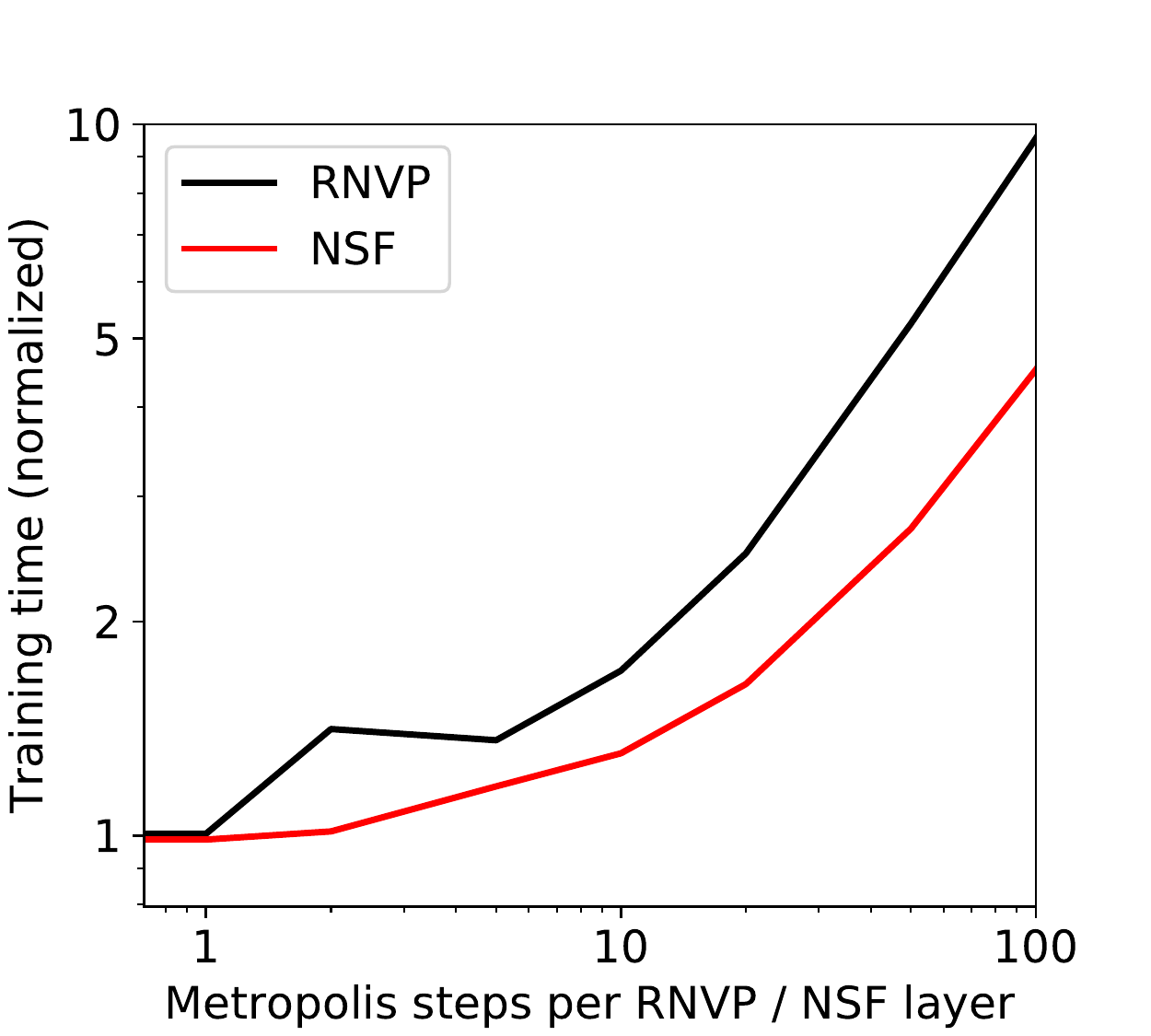}
\end{centering}
\caption{\label{fig:Images_timings}
\textbf{Computational cost of adding stochastic layers}. Time required for training SNFs of images shown in \ref{fig:Images} with a fixed number of steps, as a function of the number of stochastic layers per RNVP or NSF layer. Timings are normalized to one RNVP or NSF layer. While details of these timings depend on hyperparameters, implementation and compute platform, the main feature is that deterministic flow layers are much more computationally expensive than stochastic flow layers, and therefore a few stochastic flow layers can be added to each deterministic flow layer without significant increase in computational cost.}
\end{figure}

\end{document}